\documentclass[12pt]{article}
\usepackage[margin=1in]{geometry}

\usepackage{wrapfig}

\usepackage[utf8]{inputenc} 
\usepackage[T1]{fontenc}    
\usepackage{hyperref}       
\usepackage{url}            
\usepackage{booktabs, multicol, multirow}       
\usepackage{amsmath}
\usepackage{amsthm}
\usepackage{amsfonts}       
\usepackage{amsmath}
\usepackage{xfrac}
\usepackage{textcomp}

\usepackage{caption}
\usepackage{multirow}
\usepackage{boldline}
\usepackage{hhline}
\usepackage{enumitem}

\usepackage[numbers]{natbib}  

\usepackage{thm-restate}
\usepackage{graphicx}
\usepackage{subfig}
\usepackage{color}

\usepackage{nicefrac}       
\usepackage{microtype}      
\usepackage[toc,page]{appendix}
\usepackage{booktabs}

\usepackage{upgreek,rotating}
\newcolumntype{?}{!{\vrule width 1pt}}

\usepackage{thm-restate}
\usepackage{graphicx}
\usepackage[margin=1in]{geometry}
\usepackage{array}
    \newcolumntype{P}[1]{>{\centering\arraybackslash}p{#1}}
    \newcolumntype{M}[1]{>{\centering\arraybackslash}m{#1}}

\usepackage{caption}
\usepackage{algorithm}
\usepackage{algorithmic}

\usepackage[group-separator={,}]
{siunitx}

\newcommand{\faredust}[1]{\texttt{FARe-DUST}}
\newcommand{\feast}[1]{\texttt{FeAST-on-MSG}}
\newcommand{\fedavg}[1]{\texttt{FedAvg}}
\newcommand{\fedadam}[1]{\texttt{FedAdam}}
\newcommand{\fedbuff}[1]{\texttt{FedBuff}}
\newcommand{\pe}[1]{\texttt{per-example}}
\newcommand{\pdpe}[1]{\texttt{per-domain per-example}}

\newtheorem*{lemma*}{Lemma}

\newtheorem*{theorem*}{Theorem}
\theoremstyle{definition}

\theoremstyle{definition}

\theoremstyle{definition}

\theoremstyle{definition}

\newtheorem*{claim*}{Claim}

\title{Learning from straggler clients in federated learning}

\author{
   Andrew Hard$^{\star *}$ \quad Antonious M. Girgis$^{\dagger}$\thanks{Equal contribution} \quad Ehsan Amid$^\ddagger$\\ \quad Sean Augenstein$^\star$ \quad Lara McConnaughey$^\star$ \quad Rajiv Mathews$^\star$ \quad Rohan Anil$^\ddagger$\\
 {\small $^\star$Google Research\quad $^\dagger$University of California, Los Angeles\quad $^\ddagger$Google DeepMind}
}
\allowdisplaybreaks

\date{}

\begin{document}

\maketitle
\begin{abstract}
  How well do existing federated learning algorithms learn from client devices that return model updates with a significant time delay? Is it even possible to learn effectively from clients that report back minutes, hours, or days after being scheduled? We answer these questions by developing Monte Carlo simulations of client latency that are guided by real-world applications. We study synchronous optimization algorithms like \fedavg{} and \fedadam{} as well as the asynchronous \fedbuff{} algorithm, and observe that all these existing approaches struggle to learn from severely delayed clients. To improve upon this situation, we experiment with modifications, including distillation regularization and exponential moving averages of model weights. Finally, we introduce two new algorithms, \faredust{} and \feast{}, based on distillation and averaging, respectively. Experiments with the EMNIST, CIFAR-100, and StackOverflow benchmark federated learning tasks demonstrate that our new algorithms outperform existing ones in terms of accuracy for straggler clients, while also providing better trade-offs between training time and total accuracy.
\end{abstract}

\section{Introduction}

Cross-device federated learning (FL)~\cite{mcmahan2017communication} is a distributed computation paradigm in which a central server trains a global model by learning from data distributed across many client devices. Data heterogeneity and system heterogeneity across clients differentiate cross-device FL from other distributed optimization settings. Data heterogeneity refers to the fact that the data are not independently and identically distributed across clients (non-IID). System heterogeneity results from the many different device types that participate in training, each with different storage, communication, and computation constraints~\cite{kairouz2021advances}. This heterogeneity leads to \textit{straggler clients} in FL that take significantly more time to compute and return local model updates.

Standard FL algorithms such as \fedavg{}~\cite{mcmahan2017communication}, \fedadam{}~\cite{reddi2020adaptive}, and \texttt{FedProx}~\cite{li2020federated} utilize synchronous training rounds. In each round, a cohort of clients is sampled from a population. The clients receive a global model from the central server, update the model by processing locally cached training data, and upload the updated model back to the server. The server aggregates updated models from the clients in order to update the global model. In the synchronous setting, the server waits until all clients from the cohort (including stragglers) report back before updating the global model. The duration of each round is dictated by the slowest client in the cohort. Thus, synchronous FL algorithms can be slow to achieve a target accuracy in the presence of stragglers created by the natural heterogeneity of FL.

Production deployments perform cohort over-selection~\cite{bonawitz2019towards} to mitigate the impact of stragglers in FL. The orchestrating server selects a larger client cohort than required at the beginning of each FL round, uses the fastest clients to compute a global model update on the server, and discards the straggler client updates. Since the server doesn't have to wait for stragglers, over-selection significantly speeds up the training process and makes it robust to client dropouts. However, the final global model may be biased against stragglers.

Asynchronous optimization has been studied previously in server-based distributed settings~\cite{nedic2001distributed,mania2015perturbed,agarwal2011distributed,mishchenko2022asynchronous} as well as federated settings~\cite{xie2019asynchronous,nguyen2022federated,huba2022papaya,so2021secure}. Asynchronous FL abandons the notion of the synchronized round: the server schedules training on multiple clients in parallel and aggregates client updates as they are received. The decoupling of client computations and server updates speeds up training~\cite{nguyen2022federated}. Convergence rates for asynchronous FL are proportional to the maximum client delay~\cite{koloskova2022sharper,nguyen2022federated,stich2019error}, though convergence is not always stable empirically.

Why do we care if global models are biased against straggler clients? In real-world FL, device characteristics (RAM, connection speed) may be correlated with socioeconomic factors. Failing to learn from straggler clients could result in models that perform poorly for certain groups within a population. In the hypothetical case of federated speech models, existing straggler mitigation techniques could result in higher word error rates for some dialects.

\vspace{-0.2cm}
\subsection{Summary of contributions} \label{sec:contributions}

The goal of this work is to learn more effectively from straggler clients in FL, while maintaining competitive trade-offs for training time and overall accuracy. Our contributions are listed below.

\vspace{-0.1cm}
\begin{itemize}
\setlength\itemsep{0.0mm}
    \item We construct a realistic Monte Carlo (MC) model for simulating client latency, based on observations of real-world applications of FL.
    \item We propose different client latency distributions in order to understand how FL algorithms utilize additional training examples and novel data domains from straggler clients.
    \item We measure the performance of existing algorithms such as \fedavg{}, \fedadam{}, and \fedbuff{} in simulations with straggler clients, and identify improvements including exponential moving averages (EMA), knowledge distillation regularization, and proximal terms that improve the accuracy-training time trade-off.
    \item We propose two novel algorithms, \faredust{} and \feast{}, that are based on the concepts of knowledge distillation and EMA, respectively. These algorithms are capable of learning more from straggler clients with extreme time delays, while also training quickly and maintaining excellent overall accuracy.
\end{itemize}
\vspace{-0.1cm}

\subsection{Related work}

Stragglers are a common problem for distributed learning. Prior works have explored mitigation based on data replication and coding theory~\cite{wang2015using,karakus2017straggler,ananthanarayanan2013effective}. These techniques are incompatible with FL because the central server (by design) cannot access client data~\cite{kairouz2021advances}. Standard FL algorithms such as \fedavg{}~\cite{mcmahan2017communication} and its variants~\cite{reddi2020adaptive,li2020federated,hsu2019measuring} utilize synchronous server updates that can be delayed by stragglers. \texttt{FedProx}~\cite{li2020federated} proposed assigning different local epochs for each client depending on their local computation speed. This approach does not account for communication latency, a significant source of delay. Cohort over-selection~\cite{bonawitz2019towards} can mitigate the delay caused by stragglers in synchronous FL, but it leads to under-representation of stragglers in the global model~\cite{huba2022papaya}. Finally, \texttt{InclusiveFL}~\cite{Liu2022NoOL} addresses the straggler under-representation issue by scaling the trainable model size according to the available client resources.

\citet{xie2019asynchronous,koloskova2022sharper}, and \citet{tsitsiklis1986distributed} studied a form of asynchronous stochastic gradient descent (async-SGD) in which the server updates the global model upon receipt of each client update. Methods have been proposed to down-weight straggler updates based on delay~\cite{wang2022asyncfeded,xie2019asynchronous,zheng2017asynchronous}. While such methods can improve total accuracy and training speed, they still induce a bias against straggler clients by down-scaling or dropping extremely delayed updates. \fedbuff{}~\cite{nguyen2022federated,huba2022papaya} eschews down-weighting and aggregates client updates in a buffer before updating the global model.

Probabilistic wall clock training time models have been explored previously for distributed ML~\cite{lee2017speeding} and FL~\cite{reisizadeh2022straggler, charles2021large, so2021secure}. Prior works mostly focus on a shifted-exponential parameterization of latency and assume that runtime is proportional to the number of steps of SGD. All computation nodes feature the same latency distribution, and stragglers are distinguished only by having more training points.

Global model performance for straggler clients has also been explored in multiple contexts. \citet{huba2022papaya} showed that \fedavg{} with over-selection biased the training distribution and led to worse performance for clients with many examples, while \fedbuff{} did not. \citet{charles2021large} explored fairness via the impact of cohort size on accuracy for individual client datasets.

Scenarios in which key data domains exist predominantly on straggler clients have not been explored thoroughly. Such cases are important for real-world  FL, since stragglers often correspond to older and cheaper devices and may be correlated with socioeconomic factors. Excluding stragglers may create models that are biased and not representative of the entire population. Our work explores problems in which straggler clients feature unique data domains that are not present on fast clients.

\section{Algorithms} \label{sec:algorithms}

Let $f(w, x)$ denote the loss of model $w$ on example $x$. We considered the FL framework for solving the optimization problem: $\min_{w\in\mathbb{R}^d}\frac{1}{m}\sum F_i(w),$ where $m$ is the total number of clients and $F_i(w) = \mathbb{E}_{x \sim \mathcal{D}_i}[f(w, x)]$ is the loss function of the $i$th client, with local dataset $\mathcal{D}_i$.

We studied the performance of multiple FL algorithms for straggler clients. A few canonical baseline algorithms were considered, as well as modified versions of these baselines that were intended to learn more from stragglers. Finally, we developed two entirely new algorithms, \faredust{} and \feast{}, that were designed to learn from straggler clients with arbitrary delays.

\subsection{Baseline federated algorithms and modifications} \label{sec:baseline_algorithms}

\fedavg{}~\cite{mcmahan2017communication} and \fedadam{}~\cite{reddi2020adaptive} were used as synchronous FL baselines. Both algorithms were trained with and without cohort over-selection. \fedbuff{}~\cite{nguyen2022federated} was chosen as an asynchronous FL baseline.

Since the baseline algorithms were originally designed to maximize performance across an entire client population, we tested various modifications that would enable them to learn more from straggler clients. For the synchronous algorithms, we experimented with \textit{time-limited client computations}. In this technique, clients computed as many local steps of SGD as they could before reaching a maximum training time limit (not step limit). We chose to set time limits to the population-wide 75th percentile on-device computation times for one local training epoch. The time limit did not apply to the communication latency. While this modification still led to more training steps on fast client data, it ensured that straggler clients could report back at least partial training results quickly.

\fedbuff{} was unstable in preliminary experiments with extremely delayed clients, so we tested multiple techniques to smooth the convergence. The first was \textit{distillation regularization}~\cite{hinton2015distilling, anil2018large}, in which the most recent global model weights were sent to each client and used as a teacher network for distillation during local training. The teacher loss was interpolated with the supervised loss function using a small regularization factor. The second \fedbuff{} addition we explored was adding a \textit{proximal term} to the client loss function~\cite{li2020federated}. This regularization helped reduce client drift~\cite{Shoham2019OvercomingFI} for stragglers. Third, we tested \textit{exponential moving averages} (EMA) of model weights~\cite{polyak1992acceleration}. EMA models have been shown to generalize better than weights from individual time steps. For asynchronous algorithms, EMA might smooth out vast differences between time steps. Finally, we experimented with applying Adam instead of SGD to the \fedbuff{} server update step, since adaptive optimizers significantly improve synchronous FL~\cite{reddi2020adaptive}.


\subsection{\faredust{} algorithm} \label{sec:fare_dust}

Our first algorithm, \faredust{} (Federated Asynchronous Regularization with Distillation Using Stale Teachers), incorporates stale updates from straggler FL clients into a global model using co-distillation~\cite{anil2018large}. The key insight from prior co-distillation works is that students can benefit from stale, partially trained teacher networks.

Our algorithm builds upon \fedavg{}. At each round $t \in T$, the server samples a set $\mathcal{C}_t$ of size $B_t = \vert \mathcal{C}_t \vert$ clients uniformly at random from the population. The global model is updated as soon as the server receives the $B<B_t$ local updates from the fastest subset of clients, $\mathcal{C}_{t}^{\text{fast}}$. The fast client weight deltas ($\Delta_t^c$ for $c \in \mathcal{C}_{t}^{\text{fast}}$) are summed to produce $\Delta_t$, which is applied as a pseudo-gradient to the global model\footnote{We show the SGD update for simplicity, though other server optimizers (e.g., Adagrad, Adam, and Yogi) may be utilized in FL.}: $w_{t+1} = w_{t} - \frac{\eta_g}{B} \Delta_t$. So far, this is equivalent to \fedavg{}.

\faredust{} diverges from \fedavg{} by storing the past $k$ summed \emph{historical} model deltas ($\mathsf{Deltas}=\{\Delta_{t-k},\ldots,\Delta_{t}\}$). When straggler clients from round $t'$ (for $t-k \leq t' < t$) belatedly report back to the server, their updates are summed into a corresponding historical model delta $\Delta_{t'}$. Thus, $\Delta_{t'}$ includes updates from stragglers as well as fast clients. These weight deltas are subsequently used to create \emph{teachers} for distillation on clients.

Teacher networks are constructed by uniformly randomly sampling a historical model delta $\Delta_\text{teacher}$ from $\mathsf{Deltas}$ and applying it to the \emph{latest} model weights: $w_{\text{teacher}} = w_t - (\eta_g / B_{\text{teacher}}) \Delta_{\text{teacher}}$. Stragglers thus contribute to the historical model deltas, which create the teacher networks that distill straggler knowledge into subsequent cohorts of clients. As computation input, each sampled client receives the most recent global model, $w_t$, as well as one teacher network, $w_{\text{teacher}}$. A different teacher is sampled for every client in every round.

Clients in \faredust{} run multiple iterations of SGD to update the local model. For a given client $i$, the client loss at iteration $t$ is the sum of two functions: the local loss function $F_i(w_{t}^{i})$ from training on the local client data, and the distillation loss $\psi_i(w_{t}^{i}, w_{\text{teacher}})$ for the historical teacher model $w_{\text{teacher}}$ logits and student outputs on the local dataset $\mathcal{D}_i$. Local update step $t$ for client $i$ is therefore given by

\[
w_{t+1}^{i} = w_{t}^{i} - \eta_l \nabla_{w_{t}^{i}} \left( F_i ( w_{t}^{i} ) + \rho\, \psi_i ( w_{t}^{i}, w_{\text{teacher}} ) \right),
\]

where $\rho \geq 0$ governs the distillation regularization strength. Since the teacher model $w_{\text{teacher}}$ has updates from stragglers, the new global model can learn from the stale updates of the stragglers.   

See Algorithm~\ref{algo:fare_dust_server} in Appendix~\ref{sec:appendix_algorithms} for the complete \faredust{} procedure. At the end of training, the EMA of the global model weights is used for inference.

\subsection{\feast{} algorithm} \label{sec:feast_on_msg}
\label{sec:feastonmsg}

Our second FL algorithm is \feast{} (Federated Asynchronous Straggler Training on Mismatched and Stale Gradients). The core idea is to transfer information from early round stragglers to later rounds via an \textit{auxiliary model}. \feast{} builds on EMA~\cite{polyak1992acceleration}, and the fact that weight averaging over many steps (including stale steps) improves performance.

Let $\tau_{\max}$ denote the maximum amount of time that we allow the straggler clients to report back their local updates at each round. At the server, we maintain the global model weights $w_t$ along with \emph{auxiliary model weights} $a_t$ at round $t$. As in \fedavg{} and \faredust{}, we update the global weights to get $w_{t+1}$ using updates from the $B \leq B_t$ fastest clients $\mathcal{C}_{t}^{\text{fast}}$ that report back to the server. The global weights $w_{t+1}$ are used to start the next round of FL. However, we keep track of the global model weights $w_t$ and gradient $\Delta_t$ for the duration of $\tau_{\max}$. In the meantime, we accumulate any additional gradient from stragglers at round $t$ within the $\tau_{\max}$ time interval. Within $\tau_{\max}$, we combine the model differences from the straggler clients $\mathcal{C}_t^{\text{slow}} \subseteq \mathcal{C}_t \setminus \mathcal{C}_t^{\text{fast}}$ at round $t$ that successfully report back with the model difference from the fast clients $\Delta_t$ to form 
\begin{equation}
  \label{eq:delta-plus}
  \Delta^{+}_t = \Delta_t + \sum_{c \in \mathcal{C}_t^{\text{slow}}} \Delta^c_t.
\end{equation}


We use $\Delta^{+}_t$ to form $w_{t+1}^+$, based on $w_t$: $w_{t+1}^+ = w_{t} - ( \eta_g / B^{+}_{t} ) \Delta^{+}_t$, where $\eta_g$ is the global model learning rate and $B^+_t = \vert\mathcal{C}_t^{\text{fast}}\vert + \vert\mathcal{C}_t^{\text{slow}}\vert$.

In general, we expect $w_{t+1}^+$ to correspond to a slightly improved version of $w_{t+1}$, as we use a larger client cohort for the update step. However, we cannot replace the global model weights for the next round $w_{t+1}^+$, as the client updates at round $t+1$ are already calculated using $w_{t+1}$. While the error due to the mismatched weights and client updates may not be large for a single round, the accumulated error over multiple rounds can significantly deteriorate the performance. Therefore, we utilize the auxiliary model weights $a_t$ as a form of \emph{weight averaging} to reduce the effect of mismatched weights. Specifically, we write the auxiliary model weights update as
\begin{equation}
  \label{eq:aux_update}
  a_{t+1} = \beta (a_t - \frac{\eta_a}{B^+_t} \Delta^{+}_t) + (1 - \beta) w^+_{t+1},
\end{equation}
where $\eta_a > 0$ is the auxiliary model learning rate and $\beta \in (0, 1)$ denotes the EMA decay. The auxiliary model update in Equation~\eqref{eq:aux_update} has interesting interpretations based on the value of $\eta_a$. For $\eta_a=0$ (and assuming $a_0 = w_0^+ = w_0$), the auxiliary model update reduces to an EMA of the past models, with a decay factor of $\beta$: $a_{t+1} = \beta a_t + (1 - \beta) w^+_{t+1}$. Alternatively, when $\eta_a = \eta_g$, the updated auxiliary model is formed via a gradient step on a weighted average of the current auxiliary and global models using the aggregated model differences $\Delta^{+}_t$: $a_{t+1} = \big(\beta\,a_t + (1 - \beta)\, w_t\big) - \frac{\eta_g}{B^+_t}\,\Delta^{+}_t\,$. Algorithm~\ref{algo:feast_on_msg_server} in Appendix~\ref{sec:appendix_algorithms} provides a complete description of \feast{}.

\section{Client latency modeling} \label{sec:latency}

This section describes our MC models of client latency. Realistic client latency simulations are essential for measuring how well FL algorithms learn from stragglers. Observations from production FL systems are provided in Appendix~\ref{sec:appendix_obs_latency}.

\subsection{Components of client latency} \label{sec:latency_model}

\citet{wang2021field} provided a framework for modeling FL round and client execution time. The total run time is a sum of server-client communication time and client computation time: $T_{\text{round}} = T_{\text{communication}} + T_{\text{computation}}$, where $T_{\text{computation}} = C + (T_{\text{per-example}} \cdot N_{\text{examples}})$. $C$ is a system overhead constant for clients, and includes the start-up and turn-down time associated with the client computations. It does not scale with the number of training examples. $T_{\text{per-example}}$ is the time required to process each local training step on clients, and $T_{\text{communication}}$ is the communication time, including both the client download time and the server upload time.





\subsection{Monte Carlo latency simulation} \label{sec:latency_mc}

We directly modeled distributions of $C$, $T_{\text{per-example}}$, and $T_{\text{communication}}$ with log-normal functions. Download and upload latency were not modeled separately, since the two are strongly correlated in practice. $N_{\text{examples}}$ corresponded to the simulated client dataset size $\vert \mathcal{D}_i \vert$.


Client latencies were randomly generated via Monte Carlo techniques during FL simulations. Each client was assigned a particular $\mu$ and $\sigma$ for each latency parameter at the beginning of training. During each FL round, clients in the cohort randomly sampled latency values from their parameter distributions. Thus, clients sampled different latency values each time they participated in training.

\subsection{Latency models} \label{sec:latency_models}

In order to better understand how straggler clients affect the training time and performance of various FL algorithms, we explored two client latency models: \pe{} and \pdpe{} client latency. In particular, we were interested in whether algorithms that learned more from stragglers benefited more from additional data volume, additional data domains, or both.

The \pe{} client latency model is a stochastic model in which client computation times scale with the number of examples in the local client data. Latency factors were sampled from the MC model described in Section~\ref{sec:latency_mc} for each client during each round of FL. Specific parameterizations of the latency factors are summarized in Table~\ref{table:latency_params_pe}. This setting was useful for exploring the impact of the additional data available on straggler clients, since stragglers were only distinguished by the amount of local data and not data domain.

\begin{table}
  \caption{Latency factor parameterizations for all clients in the \pe{} client latency model.}
  \label{table:latency_params_pe}
  \vskip 0.15in
  \centering
  \begin{tabular}{ll}
    \toprule
    Latency factor & Parameterization \\
    \midrule 
    $T_{\text{communication}}$ & $\textrm{Lognormal}(\mu=2.7, \sigma=1.0)$    \\
    $T_{\text{per-example}}$   & $\textrm{Lognormal}(\mu=-1.6, \sigma=0.5)$   \\
    $C$                        & $\textrm{Lognormal}(\mu=3.0, \sigma=0.3)$    \\
    \bottomrule
  \end{tabular}
\end{table}

The \pdpe{} client latency model is a stochastic model in which clients with examples from specific data domains are assigned significantly higher latencies. This model builds on the \pe{} model by scaling client computation times with the number of client examples. However, certain clients (pre-selected according to data domain) were assigned distributions of $T_{\text{communication}}$, $T_{\text{per-example}}$, and $C$ with higher median times and longer tails. Parameterizations of the latency factors for standard clients and straggler clients are summarized in Table~\ref{table:latency_params_pdpe}. The differences between the data domains of standard clients and straggler clients in this setting enabled us to explore how FL algorithms learned from novel data domains found only on straggler clients.

\begin{table}
  \caption{Latency factor parameterizations for the \pdpe{} client latency model. Larger $\mu$ values were used for the straggler clients, resulting in higher medians and longer tails.}
  \label{table:latency_params_pdpe}
  \vskip 0.15in
  \centering
  \begin{tabular}{lll}
    \toprule
    & \multicolumn{2}{c}{Parameterization} \\
    \cmidrule(r){2-3}
    Latency factor & Standard clients & Straggler clients \\
    \midrule 
    $T_{\text{communication}}$ & $\textrm{Lognormal}(\mu=2.7, \sigma=1.0)$  & $\textrm{Lognormal}(\mu=3.7, \sigma=1.0)$    \\
    $T_{\text{per-example}}$   & $\textrm{Lognormal}(\mu=-2.0, \sigma=0.2)$ & $\textrm{Lognormal}(\mu=-1.0, \sigma=0.5)$  \\
    $C$                        & $\textrm{Lognormal}(\mu=3.0, \sigma=0.3)$  & $\textrm{Lognormal}(\mu=3.5, \sigma=0.3)$ \\
    \bottomrule
  \end{tabular}
\end{table}

\section{Datasets and tasks} \label{sec:data_and_tasks}

\subsection{Federated tasks} \label{sec:fl_tasks}

Experiments were conducted with three popular FL benchmark tasks: two dense gradient tasks (EMNIST~\cite{cohen2017emnist} and CIFAR-100~\cite{krizhevsky2009learning} classification), and one sparse gradient task (StackOverflow~\cite{stackoverflow2019} next-word prediction).

The EMNIST training data consists of handwritten characters converted into a $28 \times 28$ pixel image format. To create a simulated non-IID FL dataset, EMNIST examples were partitioned into client clusters based on the writer. We then selected a subset of EMNIST clients to represent stragglers. First, we defined the digits 0-4 as \textbf{straggler classes}. Next, we found the 800 clients with the most examples from those straggler classes and defined them as the \textbf{straggler clients}. Finally, all examples from the straggler classes were removed from non-straggler clients, which were referred to as \textbf{standard clients}. Thus, training examples from straggler classes resided exclusively in the 800 straggler client data partitions. Overall, there were 3,400 simulated FL clients for EMNIST (2,600 standard and 800 straggler clients).


The CIFAR-100 data contains 50,000 RGB images of size $32 \times 32 \times 3$ from 100 classes. We used the non-IID partitioning of the CIFAR-100 dataset proposed in \cite{reddi2020adaptive}, which created 500 clients with 100 examples each, using a two-step LDA process over the coarse and fine labels. Similar to the procedure for generating straggler clients for EMNIST, we first defined the ``vehicles 1'' and ``vehicles 2'' coarse classes as the straggler classes. As with EMNIST, we selected the 125 clients with the most straggler class examples as the straggler clients. Any examples from the straggler classes were then removed from the 375 remaining standard clients.

StackOverflow training data consists of textual question-and-answer posts to the eponymous website along with associated subject tags. Data were partitioned into simulated FL clients based on user ID. Examples with the ``math'' and ``c + +'' tags were chosen as the straggler classes for StackOverflow. All clients with at least one training example with one of these tags were labeled as straggler clients. This procedure resulted in a dataset consisting of 342,477 clients (248,595 standard and 93,882 straggler clients). Questions and answers with the straggler classes (``math'' and ``c + +'' tags) were found exclusively on the straggler clients.

We used the same model architectures as in~\citet{reddi2020adaptive} for each task: a CNN model~\cite{1315150, 5459469, Krizhevsky2010ConvolutionalDB} for EMNIST, a ResNet-18 model~\cite{He2015DeepRL} for CIFAR-100, and an LSTM~\cite{hochreiter1997long} for StackOverflow.

\subsection{Data and latency models} \label{sec:latencies_with_datasets}

The \pe{} and \pdpe{} client latency models described in Section~\ref{sec:latency_models} were explored in conjunction with the 3 FL tasks. Latency statistics sampled from one epoch over the EMINST dataset are shown in Table~\ref{table:emnist_percentiles} for both the \pe{} and \pdpe{} client latency models. With the \pe{} scenario, straggler and standard client latencies were quite similar across percentiles. For the \pdpe{} model, however, straggler clients were $3\times$ slower than standard clients across all percentiles. The median straggler was slower than the 99th percentile standard client.


\begin{table}
  \caption{Simulated client latencies with the EMNIST dataset, using the \pe{} latency model (left) and \pdpe{} latency model (right).}
  \label{table:emnist_percentiles}
  \vskip 0.15in
  \centering
  \begin{tabular}{lcccccc}
    \toprule
    & \multicolumn{3}{c}{\pe{}} & \multicolumn{3}{c}{\pdpe{}} \\
    \cmidrule(r){2-4} \cmidrule(r){5-7}
    & 50\% [s] & 95\% [s] & 99\% [s] & 50\% [s] & 95\% [s] & 99\% [s] \\
    \midrule 
    Standard clients  & 69.89 & 151.7 & 223.4 & 59.35 & 117.3 & 194.3 \\
    Straggler clients & 77.14 & 160.5 & 223.8 & 151.5 & 323.8 & 472.0 \\
    \bottomrule
  \end{tabular}
\end{table}

Appendix~\ref{sec:appendix_latency_dist} provides distributions and statistics for each dataset and latency model combination. For the CIFAR-100 task with the \pe{} latency model, the straggler and standard clients had comparable latency distributions. With the \pdpe{} model, CIFAR-100 straggler clients were $3\times$ slower than standard clients at the 99th percentile.

\begin{figure}%
  \centering
  \subfloat[\centering \pe{} client latency]{{\includegraphics[width=0.45\linewidth]{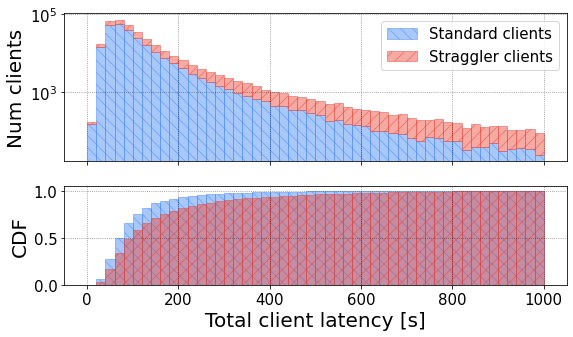} } \label{fig:stackoverflow_latency_dist_both_a}}%
  \subfloat[\centering \pdpe{} client latency]{{\includegraphics[width=0.45\linewidth]{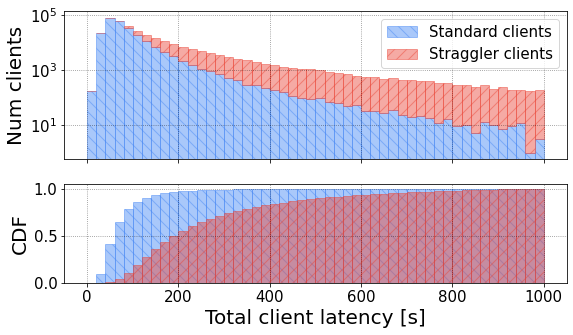} } \label{fig:stackoverflow_latency_dist_both_b}}%
  \caption{The distribution of total client latencies, in seconds, sampled for the StackOverflow dataset with the \pe{} (Figure~\ref{fig:stackoverflow_latency_dist_both_a}) and \pdpe{} client latency models (Figure~\ref{fig:stackoverflow_latency_dist_both_b}). PDFs (top) are stacked, while the CDFs (bottom) are overlaid.}%
  \label{fig:stackoverflow_latency_dist_both}%
\end{figure}

Figure~\ref{fig:stackoverflow_latency_dist_both} shows a histogram of the client latencies for StackOverflow. With the \pe{} latency model, standard and straggler clients had comparable median latencies (80.64s vs. 106.1s, respectively). Straggler clients (with ``math'' and ``c + +'' tags) had more examples than average. For the \pdpe{} latency model, the 99th percentile straggler was $10\times$ slower than the 99th percentile standard client.

\section{Experiments} \label{sec:experiments}

\fedavg{}, \fedadam{}, \fedbuff{}, \feast{}, and \faredust{} were compared on the six combinations of task and latency models described in Section~\ref{sec:latencies_with_datasets}. FL simulations were conducted using the {F}ed{JAX} framework~\cite{fedjax2021}. Training hardware consisted of a 2$\times$2 configuration of V3 TPUs~\cite{jouppitpu2017}.

\subsection{Metrics}

We defined a \textbf{straggler accuracy} metric that measured model accuracy on held-out examples belonging to the straggler data classes defined in Section~\ref{sec:fl_tasks}. This metric corresponded to classification accuracy on digits 0-4 for EMNIST and ``vehicles 1'' or ``vehicles 2'' for CIFAR-100. With StackOverflow, straggler accuracy was defined as next-word prediction accuracy for in-vocabulary tokens on posts with the ``math'' or ``c + +'' tags. The only way to improve straggler accuracy was by learning effectively from straggler clients.

We also measured total classification accuracy for EMNIST and CIFAR-100, and total in-vocabulary next-word prediction accuracy for StackOverflow. This enabled us to observe whether standard client accuracy regressed when straggler accuracy improved. The number of aggregated client updates, training steps, and total wall-clock training time were also tracked for each experiment.

\subsection{Tuning}

The objective of tuning was to maximize straggler accuracy. Total accuracy, total training time, and consistency across trials were not \textit{explicitly} optimized during the tuning process. Algorithms were tuned separately for all datasets and latency models. Default hyper-parameter values for \fedavg{} and \fedadam{} were adopted from~\cite{reddi2020adaptive}. Learning rates, buffer size, and concurrency were tuned for \fedbuff{}, while 3 parameters were tuned for both \faredust{} and \feast{}. Tuning procedures and optimal hyper-parameter values are provided in Appendix~\ref{sec:tuning}.

Training was performed for a fixed number of client updates, in order to provide fair comparisons between synchronous and asynchronous algorithms. EMNIST tasks were trained using 100,000 client updates (equivalent to 2,000 rounds of \fedavg{} with a cohort of 50 clients), CIFAR-100 tasks were trained using 80,000 client updates (4,000 rounds of \fedavg{} with a cohort of 20), and StackOverflow tasks were trained using 200,000 client updates (2,000 rounds of \fedavg{} with a cohort of 100).

\section{Results} \label{sec:results}

\begin{figure}%
  \centering
  \subfloat{{\includegraphics[width=0.48\linewidth]{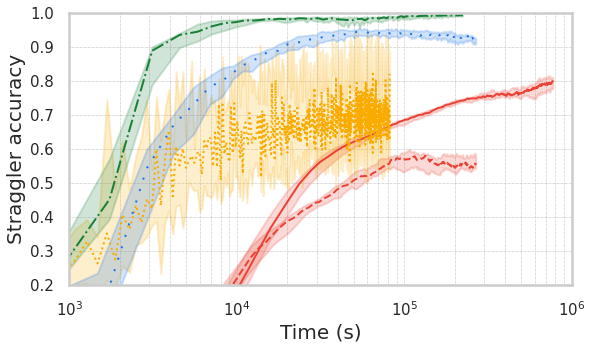} } }%
  \subfloat{{\includegraphics[width=0.48\linewidth]{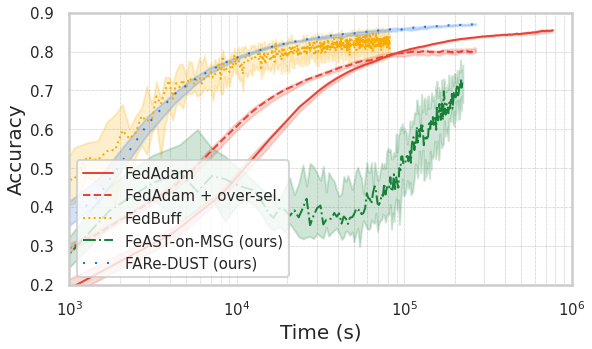} } }%
  \caption{Straggler accuracy (left) and total accuracy (right) as a function of wall clock training time for EMNIST with the \pdpe{} latency model. Solid lines and bands represent the median and 90\% confidence intervals from 10 trials.}%
  \label{fig:emnist_acc_vs_time}%
\end{figure}

Figure~\ref{fig:emnist_acc_vs_time} shows straggler accuracy and total accuracy as a function of wall-clock training time, for the EMNIST task and \pdpe{} client latency model. \feast{} and \faredust{} both converged the fastest to the highest straggler accuracy, and \faredust{} also achieved the highest total accuracy. Interestingly, \feast{} was over-tuned for straggler accuracy and had low total accuracy. \fedbuff{} trained quickly, but had poor straggler accuracy and very high-variance performance. \fedadam{} and \fedavg{} were consistent, but slow to converge and failed to reach high straggler accuracy.

\begin{figure}%
  \centering
  \subfloat{{\includegraphics[width=0.48\linewidth]{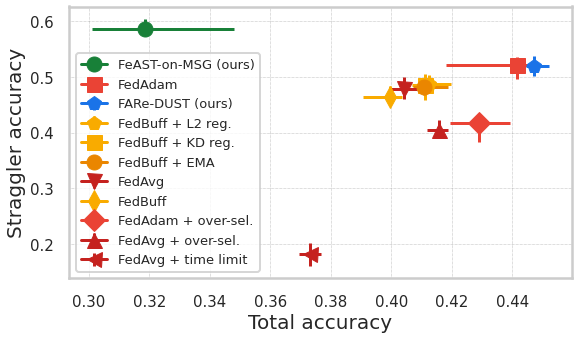} } }%
  \subfloat{{\includegraphics[width=0.48\linewidth]{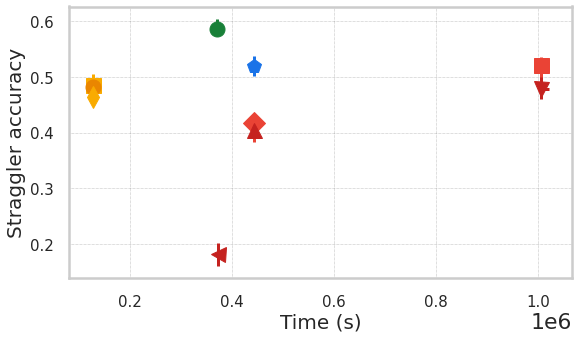} } }%
  \caption{Straggler accuracy as a function of total accuracy (left) and wall clock training time (right) for CIFAR-100 with \pdpe{} latency model. Plots show the median and 90\% CI from 10 trials for the best straggler accuracy value in each trial.}%
  \label{fig:cifar_2d}%
\end{figure}

CIFAR-100 results with the \pdpe{} client latency model are shown in Figure~\ref{fig:cifar_2d}. \faredust{} and \feast{} achieve the 1st and 3rd highest straggler accuracy values, while training more than $2.5\times$ faster than \fedadam, which had the 2nd highest straggler accuracy. While \fedbuff{} converges quickly, it fails to reach competitive total or straggler accuracy values.

Results with the \pe{} client latency model and StackOverflow task are summarized in Table~\ref{table:so_pe_main}. Values are provided for \fedavg{} and \fedadam{} with and without over-selection and with time-limited client computations. Comparisons are shown for \fedbuff{} with EMA and knowledge distillation regularization. While \fedadam{} achieved the highest straggler accuracy, it was more than $10\times$ slower than \feast{} and \faredust{}, which ranked second. \fedbuff{} once again had low straggler accuracy and total accuracy, though EMA and KD were beneficial. Interestingly, over-selection did not hurt synchronous algorithms on this task.

\begin{table}
  \caption{The straggler accuracy, total accuracy, and the total training time for each algorithm with the StackOverflow dataset and \pe{} client latency model. Median and 90\% CI values from 10 trials are quoted for each metric. Rows are sorted in descending order of straggler accuracy.}
  \label{table:so_pe_main}
  \vskip 0.15in
  \centering
  \begin{tabular}{lccccr}
    \toprule
    & \multicolumn{2}{c}{Straggler accuracy} & \multicolumn{2}{c}{Total accuracy} &                   \\
    \cmidrule(r){2-3} \cmidrule(r){4-5}

    Algorithm & Median & 95\% CI & Median & 95\% CI & Time [s] \\
    \midrule 
    FedAdam             & 25.6\% & [25.5\%, 25.7\%] & 25.4\% & [25.1\%, 25.9\%] & 3,612,004 \\
    FeAST-on-MSG        & 25.2\% & [24.6\%, 25.3\%] & 26.1\% & [25.2\%, 26.7\%] &  292,006 \\
    FARe-DUST           & 25.0\% & [24.5\%, 25.2\%] & 25.4\% & [24.3\%, 26.4\%] &  335,744 \\
    FedAdam + over-sel. & 24.9\% & [23.3\%, 25.0\%] & 25.2\% & [21.8\%, 25.9\%] &  336,140 \\
    FedBuff + EMA       & 24.3\% & [24.2\%, 24.3\%] & 24.6\% & [23.3\%, 26.0\%] &  349,178 \\
    FedBuff + KD reg.   & 24.2\% & [24.2\%, 24.3\%] & 23.9\% & [23.4\%, 25.1\%] &  350,286 \\
    FedAvg              & 23.7\% & [23.5\%, 23.8\%] & 23.6\% & [22.2\%, 24.6\%] & 3,584,260 \\
    FedBuff             & 23.3\% & [22.3\%, 23.6\%] & 23.4\% & [22.3\%, 23.7\%] &  349,257 \\
    FedAvg + time limit & 21.7\% & [21.7\%, 21.7\%] & 22.7\% & [21.3\%, 23.4\%] &  287,952 \\
    FedAvg + over-sel.  & 20.9\% & [20.8\%, 20.9\%] & 22.4\% & [21.1\%, 23.2\%] &  407,588 \\
    \bottomrule
  \end{tabular}
\end{table}

Appendix~\ref{sec:additional_observations} discusses all of the findings above in detail, and Appendix~\ref{sec:full_results} provides comprehensive numerical results for every task, latency model, and algorithm. The primary empirical observations are summarized below.

\vspace{-0.1cm}
\begin{itemize}
\setlength\itemsep{0.0mm}
    \item \faredust{} was the most consistently high-performing algorithm in terms of straggler accuracy, across tasks and latency models.
    \item \feast{} provided the highest straggler accuracy on most tasks and a good trade-off between total accuracy and time. It had a tendency to over-tune on straggler accuracy.
    \item \fedbuff{} converged faster but to lower accuracy than other algorithms on most tasks. Modifications such as EMA, distillation regularization, and adaptive optimizers improved the straggler accuracy and total accuracy for \fedbuff{}.
    \item Over-selection sped up synchronous algorithms and even boosted accuracy in the \pe{} latency scenario. However, it generally led to worse total accuracy and straggler accuracy in the \pdpe{} latency scenario.
    \item Limiting client computations by time sped up training for \fedavg{} at the cost of reduced straggler accuracy for the \pdpe{} latency scenario. Stragglers only trained for a fraction of a local epoch, while standard clients processed more than 1 epoch.
\end{itemize}
\vspace{-0.1cm}

\section{Conclusion} \label{sec:conclusion}

This work explored methods for learning from straggler clients with extreme delays in FL. Through simulation experiments with MC latency models on EMNIST, CIFAR-100, and StackOverflow, we found that our proposed algorithms --- \faredust{} and \feast{} --- significantly improved the accuracy of global models for straggler clients. Furthermore, these algorithms provided trade-offs of total accuracy and training time that were better than those offered by existing algorithms. Finally, we identified techniques such as EMA and distillation regularization that improved baselines like \fedbuff{}. Given the growth of cross-device FL applications in which device heterogeneity mirrors socioeconomic status, it is imperative that FL models are reflective of the whole population.

\bibliographystyle{unsrtnat}
\bibliography{references}

\newpage
\onecolumn
\appendix
\section{Algorithm details for \faredust{} and \feast{}}
\label{sec:appendix_algorithms}

\subsection{Motivation}

This work began with applying small modifications (EMA, knowledge distillation) to \fedbuff{} for the purpose of improving the representation of straggler clients in the learned model. These techniques are introduced in Section~\ref{sec:baseline_algorithms}, and observations are summarized in Appendix~\ref{sec:additional_observations}. Each technique provided improvements over the baseline \fedbuff{} algorithm, as summarized in Tables~\ref{table:emnist_pe_full}, \ref{table:cifar_pe_full}, \ref{table:cifar_pdpe_full}, \ref{table:so_pe_full}, and \ref{table:so_pdpe_full}. These incremental benefits motivated the development of new algorithms built around three concepts:

\vspace{-0.1cm}
\begin{enumerate}
  \setlength\itemsep{0.0mm}
  \item knowledge distillation to incorporate stale updates,
  \item EMA and model averaging to combine models trained on different data partitions,
  \item application of gradients to mismatched network weights.
\end{enumerate}
\vspace{-0.1cm}

We provide some background and motivation for each of these concepts below.

\paragraph{Distillation} Co-distillation~\cite{anil2018large} originally used distillation regularization to overcome the limitations of scaling batch size alone for parallel training on massive datasets. \citet{sodhani2021closer} hinted that co-distillation could be used to speed up federated training. It was therefore reasonable to suspect that distillation might bring data parallelism benefits to the asynchronous FL regime.

There was also evidence that distillation could transfer domain knowledge from a teacher to a student even for domains that were absent from the distillation training dataset. \citet{houyon2023online} demonstrated that distillation could be used to mitigate cyclical domain shifts. And \citet{patientconsistent9879513} showed that student models could learn classes from the soft labels of a teacher, even when those classes were not present in the student's training data.

These prior works led to our hypothesis that, \textit{if stale updates were considered as a data class of their own, knowledge distillation from a stale teacher could be used to impart knowledge of the stale class to faster clients in FL}. The strong performance of \faredust{} in experiments (high straggler accuracy, total accuracy, and reduced training latency) supports this hypothesis.

\paragraph{EMA and model averaging} Model averaging has shown promising results when combining models fine-tuned with different hyper-parameters or data subsets~\cite{wortsman2022model}, combining models trained with noisy gradients with varying data distributions~\cite{Shejwalkar2022RecyclingSI}, and as an effective way of approximating the sharpness of the loss~\cite{Du2022SharpnessAwareTF}. Prior works suggest that model averaging can result in wider optima and better generalization~\cite{Izmailov2018AveragingWL}.

A similar line of work on mode connectivity~\cite{garipov2018loss, draxler2018essentially, frankle2020linear} has shown that the optima of SGD-trained neural networks are connected by smooth curves over which accuracy and loss are nearly constant. In particular, \citet{frankle2020linear} demonstrated the existence of linear mode connectivity for checkpoints derived from a recent common ancestor checkpoint. This is quite similar to the finding of \citet{wortsman2022model}, which found that models could be averaged effectively as long as they shared a common initial checkpoint for fine-tuning.

The averaging and mode connectivity literature motivated the application of EMA and model averaging to the stale weight update problem in this work. The wider optima and better generalization provided by EMA would be beneficial to FL, since models learned on a subset of clients must generalize to unseen clients. Furthermore, the wider optima provided by EMA would likely be more amenable to averaging across multiple time steps. Linear mode connectivity suggested that extremely stale weight updates could still be averaged with more recent global models, since straggler weight updates would share a common ancestor with more recent checkpoints created by fast FL clients.

\paragraph{Mismatched gradients} The model averaging and mode connectivity works also led us to a hypothesis that \textit{it might also be possible to transfer gradients from one model to another in certain cases}. Particularly for two models with mode connectivity or wide minima, transferring gradients associated with one set of weights to another set of weights could effectively transfer knowledge about the learned data. In fact, gradient transfer has been used in \citet{augenstein2022mixed} to train models on disjoint datasets.

\paragraph{Motivation for \faredust{}} Both of our algorithms utilize straggler aggregation, in which straggler client weight updates $\Delta_{t'}$ continue to be aggregated by the server even after FL round $t$ is completed and the fast client updates $\Delta_t$ are applied to advance the global model: $w_{t+1} \gets w_{t} - \frac{\eta_g}{B} \Delta_t$. The \faredust{} algorithm also relies on the three techniques of mismatched gradients, distillation for knowledge transfer, and EMA.

\vspace{-0.1cm}
\begin{itemize}
  \setlength\itemsep{0.0mm}
  \item Mismatched gradients: teacher networks are constructed by applying the straggler weight deltas ($\Delta_{t'}$ to the \textit{most recent global model} rather than the global model which was used to compute the gradients.
  \item Distillation for knowledge transfer: the teachers formed from mismatched gradients are used for distillation regularization in subsequent training rounds. Straggler client knowledge from the mismatched gradients thus gets distilled into all clients in the current round.
  \item EMA: the final model is produced by applying EMA to the main model weights $(w_{T}, w_{T-1}, w_{T-2}, ...)$.
\end{itemize}
\vspace{-0.1cm}

\begin{figure}%
  \centering
  \subfloat[\textbf{Stale gradient accumulation}. Historical weight deltas $\Delta_{t'}, \Delta_{t+1'}, \Delta_{t+2'}, \Delta_{t+3'}, ...$ are maintained for up to $k$ prior FL rounds behind the primary weight branch. Updates from straggler clients continue to accumulate in these weight deltas even after the primary weight branch ($w_t, w_{t+1}, w_{t+2}, ...$) has advanced.]{{\includegraphics[width=0.82\linewidth]{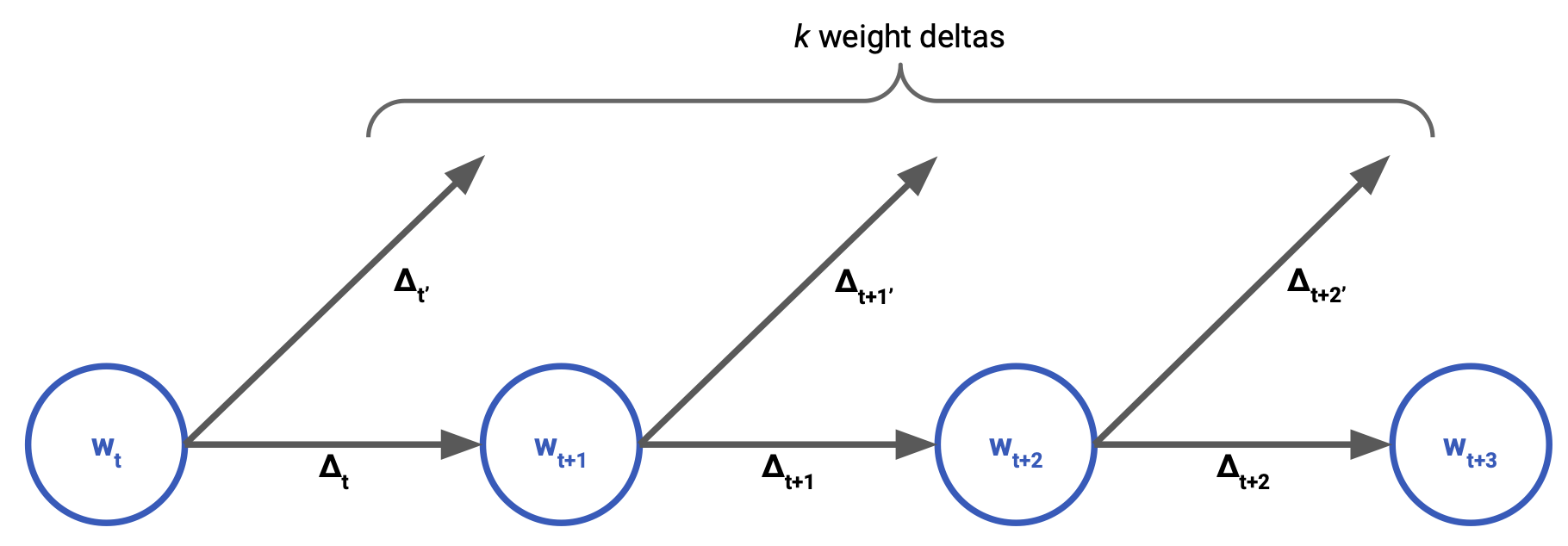}} \label{fig:faredust_diagram_a}}%
  \\
  \subfloat[\textbf{Teacher network creation}. For a given client and given FL training round, an historical weight delta is selected at random from the $k$ possible choices to construct a teacher network. In this diagram, $\Delta_{t'}$ is selected and applied to the current global model $w_{t+2}$ to form the teacher network $w_{t+1}^{+}$ via the update: $w_{t+1}^{+} = w_{t+2} - (\frac{\eta_g}{B_{t}^{+}}) \Delta_{t'}$. ]{{\includegraphics[width=0.82\linewidth]{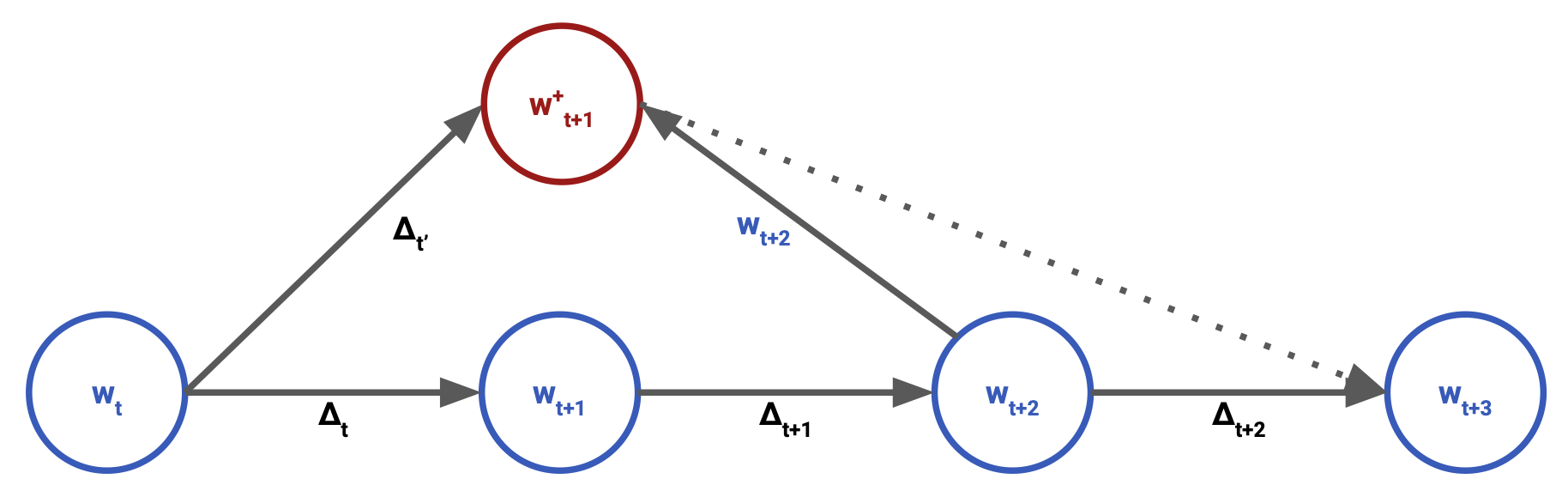}} \label{fig:faredust_diagram_b}}%
  \\
  \subfloat[\textbf{Stale teacher distillation}. Each client in each round is randomly assigned one of the possible $k$ historical teachers for distillation regularization. In this example, the initial global model $w_{t+2}$ is sent to each client along with one of the historical teachers ($w_{t}^{+}, w_{t+1}^{+}, w_{t+2}^{+}$). A new global model, $w_{t+3}$, is formed by aggregating the local client updates that were formed by training with the standard supervised and distillation regularization objectives. After $T$ rounds, $w_{T}$ (or the EMA of $w_{T}, w_{T-1}, ...$) is returned as the final model.]{{\includegraphics[width=0.82\linewidth]{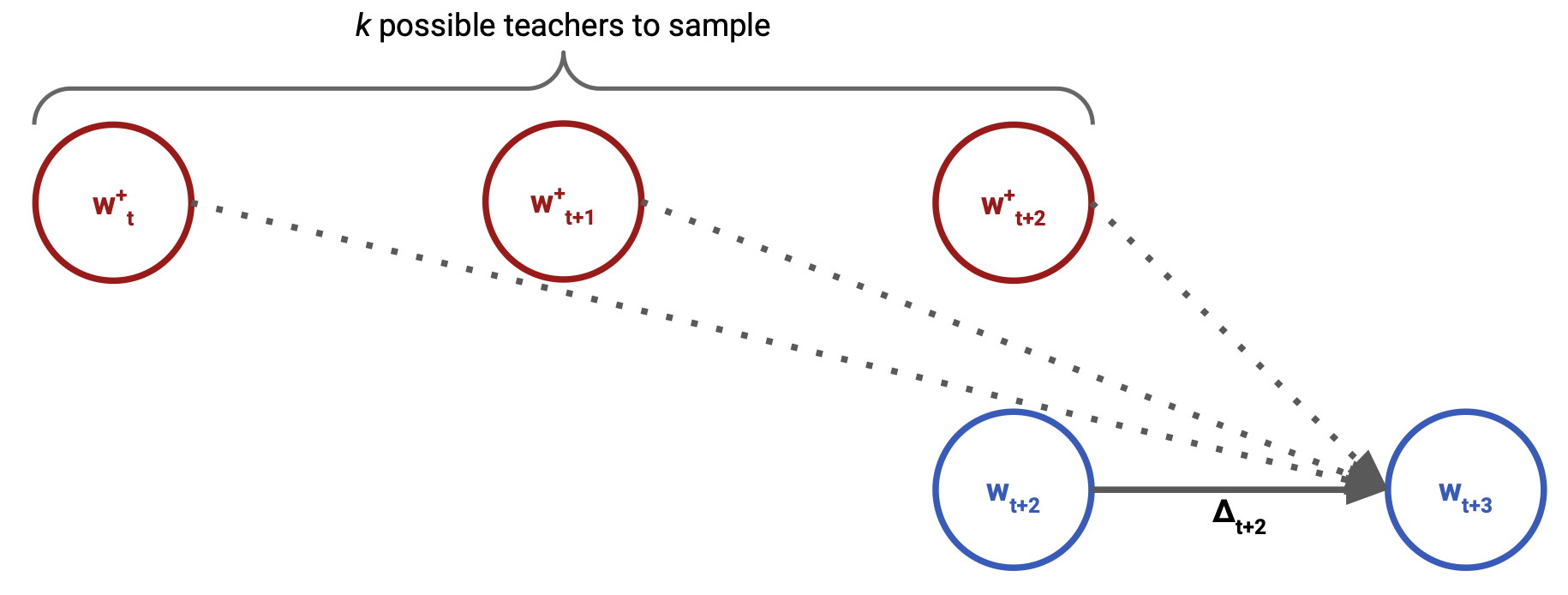}} \label{fig:faredust_diagram_c}}%
  \caption{A diagram of the three main components of \faredust{}: stale gradient accumulation (Figure~\ref{fig:faredust_diagram_a}), teacher network creation (Figure~\ref{fig:faredust_diagram_b}), and stale teacher distillation (Figure~\ref{fig:faredust_diagram_c}).}%
  \label{fig:faredust_diagram}%
\end{figure}

Figure~\ref{fig:faredust_diagram} illustrates the novel aspects of the \faredust{} algorithm.

\paragraph{Motivation for \feast{}} In addition to straggler aggregation, the \feast{} algorithm uses the techniques of mismatched gradients and EMA.

\vspace{-0.1cm}
\begin{itemize}
  \setlength\itemsep{0.0mm}
  \item Mismatched gradients: \feast{} applies the straggler weight deltas ($\Delta_{t'}$ to the auxiliary model weights $a_t$, as well as to the corresponding main global model $w_t$.
  \item EMA: the auxiliary model $a_{t+1}$ is formed using weight averaging of the straggler-updated model weights $w_{t}^{+}$ and the previous auxiliary model with the mismatched gradient applied $(a_t - \frac{\eta_a}{B_{t}^{+}} \Delta_{t}^{+})$.
\end{itemize}
\vspace{-0.1cm}

\begin{figure}%
  \centering
  \includegraphics[width=0.95\linewidth]{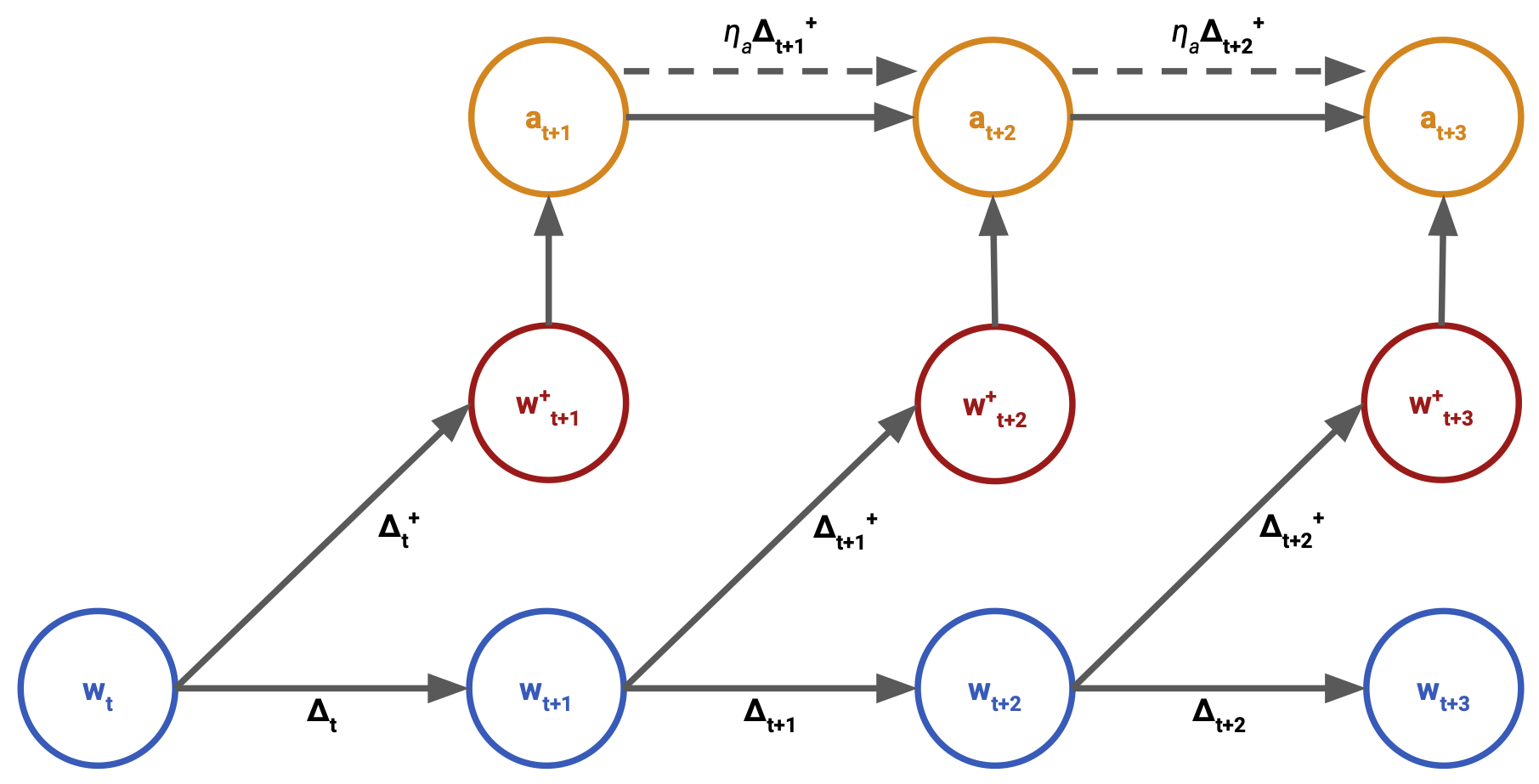}%
  \caption{A visual diagram of the \feast{} algorithm. The primary weight branch ($w_{t}, w_{t+1}, w_{t+2}, ...$) is advanced as usual for \fedavg{}. However, additional straggler updates are also allowed to accumulate for each checkpoint as they are received by the server $(\Delta_{t}^{+}, \Delta_{t+1}^{+}, \Delta_{t+2}^{+}, ...)$, up to a maximum wait time $\tau_{\textrm{max}}$. Once the maximum wait time has elapsed, an historical checkpoint is formed using the update: $w_{t+1}^{+} = w_t - (\frac{\eta_{g}}{B_{t}^{+}}) \Delta_{t}^{+}$. Historical checkpoints $w_{t+1}^{+}, w_{t+2}^{+}, w_{t+3}^{+}, ...$ are formed as the maximum wait times for each round elapse. An auxiliary branch $(a_{t+1}, a_{t+2}, a_{t+3}, ...)$ is advanced based on the historical checkpoints and stale gradients: $a_{t+1} = \beta (a_t - \frac{\eta_a}{B^+_t} \Delta^{+}_t) + (1 - \beta) w^+_{t+1}$. After a maximum of $T$ training rounds, $a_{T}$ is returned as the final model.}%
  \label{fig:feast_diagram}%
\end{figure}

Figure~\ref{fig:feast_diagram} illustrates the components of the \feast{} algorithm. Equation~\ref{eq:aux_update} also provides the exact \feast{} update formula. There are two special cases for the algorithm. First, when $\eta_a = 0$ with $a_0 = w_{0}^{+} = w_0$, the auxiliary model update is just an EMA of the past models with decay $\beta$. Second, when $\eta_a = \eta_g$, the auxiliary model update is essentially a gradient step on a weighted average of the current auxiliary and global models.

Our empirical evaluation of \feast{} validates that \textit{model averaging and mismatched gradient transfer can successfully transmit information between models trained on different data partitions}.

\subsection{Pseudo-code}

This section provides pseudo-code for the algorithms presented in Section~\ref{sec:algorithms}. The server and client computation pseudo-code for \faredust{} are shown in Algorithm~\ref{algo:fare_dust_server} and Algorithm~\ref{algo:fare_dust_client}, respectively. The \feast{} server computation pseudo-code is shown in Algorithm~\ref{algo:feast_on_msg_server}.

\begin{algorithm}[ht]
  \caption{\faredust{}: server-side}
  \label{algo:fare_dust_server}
  {\textbf{Input:}} client cohort size $B$, server learning rate $\eta_g$, client learning rate $\eta_l$, maximum number of teachers $k$, distillation strength parameter $\rho$, and EMA decay $\beta$.
  \begin{algorithmic}[1]
    \STATE \textbf{Initialize:} $w_0\in\mathbb{R}^d$, $t\gets 0$, $\mathsf{Deltas}=\lbrace \rbrace$
    \WHILE {$t\leq T$}
      \STATE Sample a subset $\mathcal{C}_t$ of size $B_t=\vert \mathcal{C}_t \vert$ clients.
      \STATE $b_t\gets 0$, $\Delta_t \gets 0$.
      \FOR {each client $c \in \mathcal{C}_t$ \textbf{in parallel}}
        \STATE Sample uniformly $\Delta_{\text{teacher}}^c$ from $\mathsf{Deltas}$.
        \STATE $w_{\text{teacher}}^c \gets w_t - \frac{\eta_g}{B_{\text{teacher}}^c} \Delta_{\text{teacher}}^c$.
        \STATE Run $\texttt{FARe-DUST}_{\text{client}}(w_t, w_{\text{teacher}}^c)$.
      \ENDFOR
      \WHILE{$b_t < B$}
        \IF {receive client $c\in\mathcal{C}_t$ update}
        \STATE $\Delta_t \gets \Delta_t + \Delta_t^{c}$,  $b_t\gets b_t+1$.
        \ENDIF
      \ENDWHILE
      \STATE $w_{t+1}\gets w_t-\frac{\eta_g}{B}\Delta_t$, $t\gets t+1$.
      \STATE $\Delta_t\gets0$, $b_t\gets 0$
      \STATE Add $\left(\Delta_{t},b_{t}\right)$ to the set $\mathsf{Deltas}$.
      \IF{receive client $c\in\mathcal{C}_{t'}, \, t' \leq t$ update}
      \STATE $\Delta_{t'}\gets \Delta_{t'} + \Delta_{t'}^{c}$, $b_{t'}\gets b_{t'}+1$. \label{line:update_teacher}
      \ENDIF
      \IF { $|\mathsf{Deltas}|> k$}
      \STATE $w_{t-k-1}\gets w_{t-k-1}-\frac{\eta_g}{b_{t-k-1}}\Delta_{t-k-1}$.
      \STATE Remove $\Delta_{t-k-1}$ from $\mathsf{Deltas}$.
      \ENDIF
    \ENDWHILE
  \end{algorithmic}
  \textbf{Output:} Global model:  $\mathsf{EMA}_{\beta}\left(w_1,\ldots, w_T\right)$.
\end{algorithm}

\begin{algorithm}[t]
  \caption{\feast{}: server-side}
  \label{algo:feast_on_msg_server}
  {\textbf{Input:}} cohort size $B$, server learning rate $\eta_g$, auxiliary learning rate $\eta_a$, client learning rate $\eta_l$, maximum wait time $\tau_{\max}$, and EMA decay $\beta$.
  \begin{algorithmic}[1]
    \STATE \textbf{Initialize:} $w_0\in\mathbb{R}^d$, $t\gets 0$, $a_0 \gets w_0$
    \WHILE {$t\leq T$}
      \STATE Sample a subset $\mathcal{C}_t$ of $B_t$ clients and run client computations.
      \STATE $b_t\gets 0$, $\Delta_t \gets 0$.
      \STATE Reset and record \texttt{elapsed\char`_time}.
      \WHILE{$b_t < B$}
        \STATE Receive client $c\in\mathcal{C}_t$ update.
        \STATE $\Delta_t \gets \Delta_t + \Delta_t^{c}$,  $b_t\gets b_t+1$.
      \ENDWHILE
      \STATE $w_{t+1}\gets w_t-\frac{\eta_g}{b_t}\Delta_t$, $t\gets t+1$.
      \STATE $\Delta_t^+ \gets \Delta_t$.
      \WHILE{\texttt{elapsed\char`_time} $\leq \tau_{\max}$}
        \STATE Receive client $c\in\mathcal{C}_t$ update.
        \STATE $\Delta^+_t \gets \Delta^+_t + \Delta_t^{c}$,  $b_t\gets b_t+1$.
      \ENDWHILE
      \STATE $w_{t+1}^+ \gets w_{t} - \frac{\eta_g}{b_t}\, \Delta^{+}_t$.
      \STATE $a_{t+1} \gets \beta\,(a_t - \frac{\eta_a}{b_t}\,\Delta^{+}_t) + (1 - \beta)\, w^+_{t+1}$.
    \ENDWHILE
  \end{algorithmic}
  \textbf{Output:} Global model: $a_T$.
\end{algorithm}

\begin{algorithm}
  \caption{$\texttt{FARe-DUST}_{\text{client}}$: client-side}
  \label{algo:fare_dust_client}
  {\textbf{Input:}} model $w$, teacher model $w_{\text{teacher}}$, local iterations $T_l$, loss function $F_i$, distillation loss $\psi$, distillation parameter $\rho$, and learning rate $\eta_l$.
  \begin{algorithmic}[1]
    \STATE \textbf{Initialize:} $w_0^{i}\gets w$.
    \FOR {$j=0,\ldots,T_l-1$}
      \STATE $w_{j+1}^{i}\gets w_{j}^{i}-\eta_l \nabla_{w_{j}^{i}}\left( F_i\left(w_{j}^{i}\right)+\rho\, \psi\left(w_{j}^{i},w_{\text{teacher}}\right)\right) $
    \ENDFOR
  \end{algorithmic}
  \textbf{Output:} Local update:  $\Delta^{i}=w_{T_l}^{i}-w$.
\end{algorithm}

\section{Additional Observations} \label{sec:additional_observations}

\paragraph{FARe-DUST} \faredust{} was the most consistently high-performing algorithm across tasks, providing an excellent trade-off between total training time, straggler accuracy, and total accuracy. In the EMNIST + \pdpe{} experiments (Figure~\ref{fig:2d_acc_emnist_pdpe_all}), \faredust{} achieved the best trade-off of straggler accuracy and total accuracy of any algorithm by a significant margin (\feast{} had better straggler accuracy, at the cost of much lower total accuracy), while also converging quickly. In experiments with CIFAR-100 and the \pe{} client latency model, \faredust{} again had the best combination of straggler accuracy and total accuracy (Figure~\ref{fig:2d_acc_cifar_pe_all}). And with the \pdpe{} client latency scenario, \faredust{} and \fedadam{} were tied for best accuracy trade-offs (Figure~\ref{fig:2d_acc_cifar_pdpe_all}). However, \faredust{} was also $2.5\times$ faster in terms of wall-clock time than \fedadam{} in this setting (Figure~\ref{fig:2d_time_cifar_pdpe_all}). Finally, on the StackOverflow task with the \pdpe{} client latency model, \faredust{} achieved the 2nd highest straggler accuracy and total accuracy (Figure~\ref{fig:2d_acc_so_pdpe_all}), but was $10\times$ faster than \fedadam{}, which had the highest straggler accuracy (Figure~\ref{fig:2d_time_so_pdpe_all}). Generally speaking, \faredust{} was faster than \fedadam{} without over-selection, and had higher total and straggler accuracy than \fedbuff{} or \fedadam{} with over-selection. \faredust{} performed particularly well in the \pdpe{} latency scenario, indicating that it was able to learn very effectively from clients that were significantly delayed.

\paragraph{FeAST-on-MSG} \feast{} also provided high straggler and total accuracy across most tasks and provided a good trade-off between training time and accuracy. Generally speaking, \feast{} was comparable in terms of total training time with \faredust{}, faster than \fedadam{} without over-selection, and had higher accuracy than \fedbuff{}. In experiments with EMNIST and the \pe{} client latency model (Table~\ref{table:emnist_pe_full}), \feast{} tied with \fedadam + over-selection for the best straggler accuracy and total accuracy while being significantly faster. For StackOverflow experiments with the \pe{} client latency scenario, \feast{} achieved the best trade-off between straggler accuracy and total accuracy. It was second in straggler accuracy only to \fedadam{} (Figure~\ref{fig:2d_acc_so_pe_all}), and was $10\times$ faster in terms of wall-clock time (Figure~\ref{fig:2d_time_so_pe_all}). And in StackOverflow experiments with the \pdpe{} client latency model, \feast{} was second only to \faredust{} in terms of the accuracy-time trade-off (Figure~\ref{fig:2d_time_so_pdpe_all}). However, \feast{} had a drawback: it was susceptible to over-tuning on straggler accuracy at the expense of total accuracy. Evidence of this can be seen in the EMNIST (Figure~\ref{fig:2d_acc_emnist_pdpe_all}) and CIFAR-100 (Figure~\ref{fig:2d_acc_cifar_pdpe_all}) experiments with the \pdpe{} client latency scenario, in which \feast{} achieved unrivaled straggler accuracies of 99.3\% and 58.6\%, respectively, while also having lower total accuracy than all other algorithms.

Large values for $\kappa = (\eta_a / \eta_g)$, which controls the auxiliary learning rate, help explain why \feast{} over-fits to straggler accuracy in these two cases (see Tables~\ref{table:feast_on_msg_params_pe} and~\ref{table:feast_on_msg_params_pdpe}). When $\kappa = 0$, \feast{} reduces to EMA on the past models. Larger values of $\kappa$, on the other hand, give greater weight to the mismatched gradient $\Delta_{t}^{+}$ applied to the auxiliary model $a_{t}$. Thus, $\kappa$ has an important role in regulating the trade-off between straggler client accuracy and non-straggler client accuracy for \feast{}.

\paragraph{FedAdam} When used with over-selection, \fedadam{} provided excellent trade-offs between straggler accuracy, total accuracy, and training time in the \pe{} client latency scenario. For example, \fedadam{} with over-selection achieved the highest straggler accuracy with the \pe{} client latency scenario for EMNIST (Table~\ref{table:emnist_pe_full}) and CIFAR-100 (Table~\ref{table:cifar_pe_full}). This result makes sense because straggler clients in this latency scenario were distinguished by data volume rather than data class. However, \fedadam{} with over-selection fared less well under the \pdpe{} client latency scenario. When using over-selection, straggler clients with straggler data were never aggregated into the global model. And without over-selection, \fedadam{} was significantly slower to converge than other algorithms. This can be observed in the \pdpe{} client latency model results with EMNIST (Table~\ref{table:emnist_pdpe_full}) and CIFAR-100 (Table~\ref{table:cifar_pdpe_full}) 

\paragraph{FedBuff} \fedbuff{} provided the fastest convergence by far on EMNIST and CIFAR-100, but yielded lower total accuracy (Figures~\ref{fig:acc_total_emnist_pdpe_all} and~\ref{fig:acc_total_cifar_pdpe_all}) and straggler accuracy (Figures~\ref{fig:acc_straggler_emnist_pdpe_all} and~\ref{fig:acc_straggler_cifar_pdpe_all}). For the StackOverflow task, \fedbuff{} was actually slower than the synchronous algorithms with over-selection (Figure~\ref{fig:acc_total_so_pe_all}). This was due to the fact that \fedbuff{} was unstable on the StackOverflow tasks when run with small buffer sizes. Tuning experiments identified a buffer size of 50 along with a max concurrency of 100 as the most stable and accurate configuration. For comparison, \fedbuff{} worked with a buffer size of 20 and max concurrency of 60 for CIFAR-100, and a buffer size of 20 with a max concurrency of 100 for EMNIST. It should also be noted that, even after tuning buffer size, max concurrency, and learning rates, the \fedbuff{} algorithm accuracy was highly variable from one round to the next in every task. We chose to explore regularization techniques for the purpose of reducing the performance variance in \fedbuff{}.

\begin{figure}%
  \centering
  \subfloat{{\includegraphics[width=0.48\linewidth]{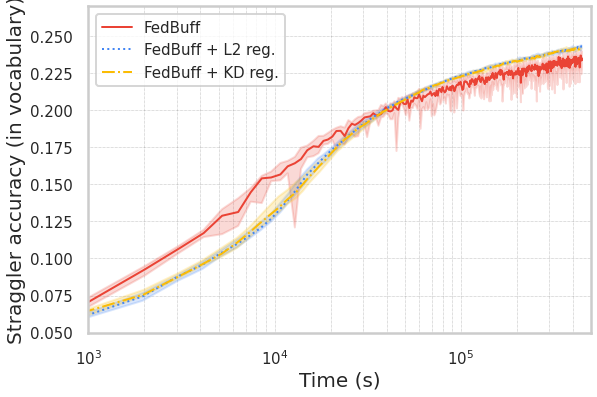}}}%
  \subfloat{{\includegraphics[width=0.48\linewidth]{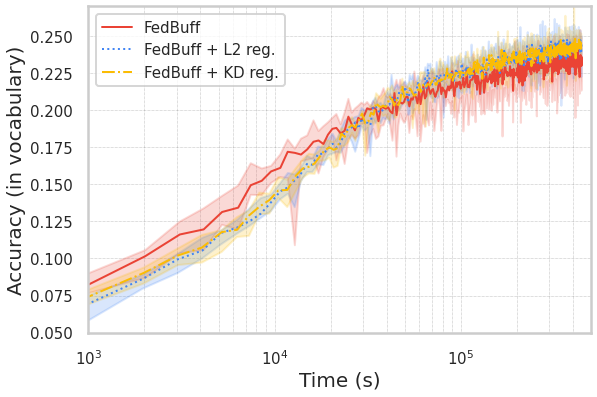}}}%
  \caption{Straggler accuracy (left) and total accuracy (right) as a function wall clock training time for StackOverflow with the \pdpe{} client latency scenario. The training curves for the \fedbuff{} algorithm are shown with and without distillation and proximal $L_2$ regularization.}%
  \label{fig:fedbuff_so_reg_comparison}%
\end{figure}

\paragraph{Regularized FedBuff} We experimented with regularized versions of the \fedbuff{} algorithm including distillation regularization and proximal $L_2$ regularization. Figure~\ref{fig:fedbuff_so_reg_comparison} compares accuracy for these variants of \fedbuff{} on the StackOverflow task with the \pdpe{} client latency model. Both forms of regularization lead to improved total accuracy and straggler accuracy. For the most part, these regularization techniques were beneficial in terms of accuracy (and reduced variance in performance) across tasks and latency models. We hypothesized that both techniques were able to reduce client drift for clients in the buffer, even when the client computations originated from different time steps.

\begin{figure}%
  \centering
  \subfloat[\centering \pe{} client latency]{{\includegraphics[width=0.48\linewidth]{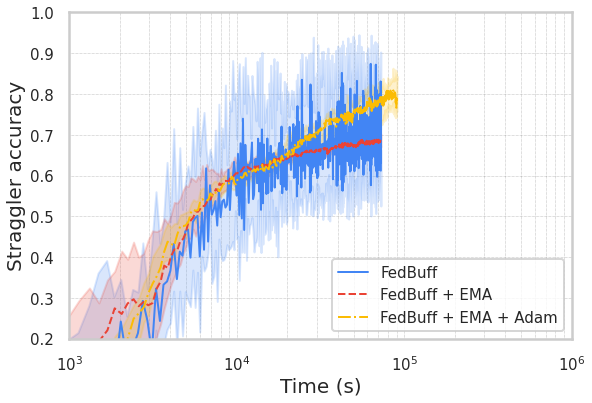}}}%
  \subfloat[\centering \pe{} client latency]{{\includegraphics[width=0.48\linewidth]{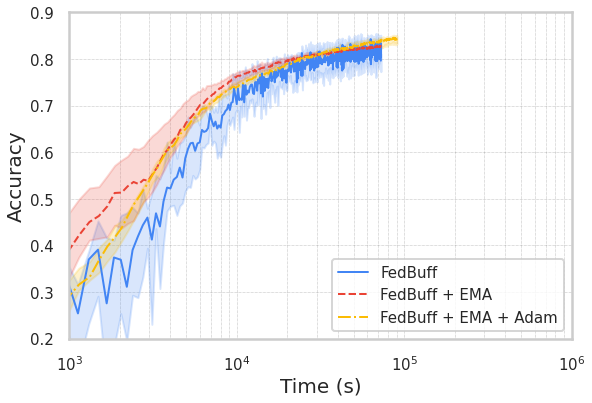}}}%
  \\
  \subfloat[\centering \pdpe{} client latency]{{\includegraphics[width=0.48\linewidth]{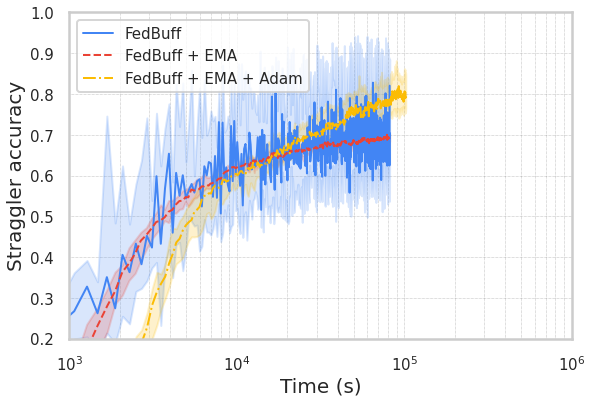}}}%
  \subfloat[\centering \pdpe{} client latency]{{\includegraphics[width=0.48\linewidth]{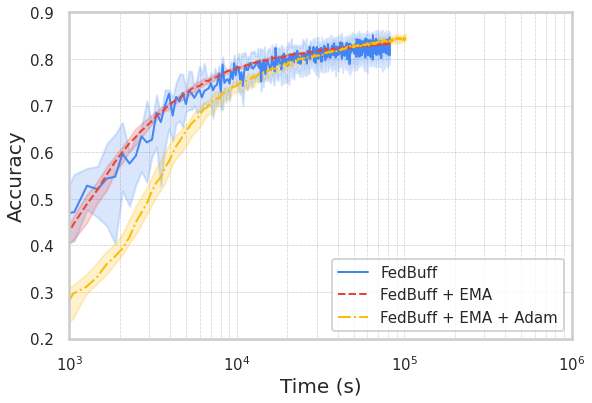}}}%
  \caption{Straggler accuracy (left) and total accuracy (right) as a function wall clock training time for EMNIST with the \pe{} client latency (top) and \pdpe{} client latency scenario (bottom). The training curves for the \fedbuff{} algorithm are shown with and without EMA and the Adam server optimizer.}%
  \label{fig:fedbuff_emnist_ema_comparison}%
\end{figure}

\paragraph{FedBuff with EMA} Figure~\ref{fig:fedbuff_emnist_ema_comparison} compares \fedbuff{} with EMA ($\beta=0.99$) and without EMA on the model weights, for the EMNIST task. While the median straggler accuracy and total accuracy were comparable in both latency scenarios, the variance of the performance was significantly lower when using EMA.

\paragraph{Adaptive FedBuff} Figure~\ref{fig:fedbuff_emnist_ema_comparison} also compares \fedbuff{} with SGD as the server optimizer (default) and with Adam as the server optimizer. The results show that Adam significantly improved the straggler accuracy and total accuracy of \fedbuff{}. Unfortunately, we did not have time to provide results with Adam as the server optimizer for all tasks in this work. However, the results suggested that, as with synchronous FL and \fedadam{}, asynchronous FL algorithms might benefit significantly from adaptive server optimizers.

\begin{figure}%
  \centering
  \subfloat[\centering \pe{} client latency]{{\includegraphics[width=0.48\linewidth]{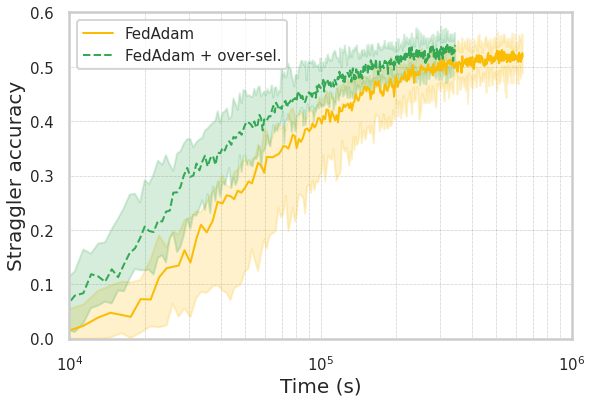}} \label{fig:fedadam_cifar_oversel_comparison_a}}%
  \subfloat[\centering \pe{} client latency]{{\includegraphics[width=0.48\linewidth]{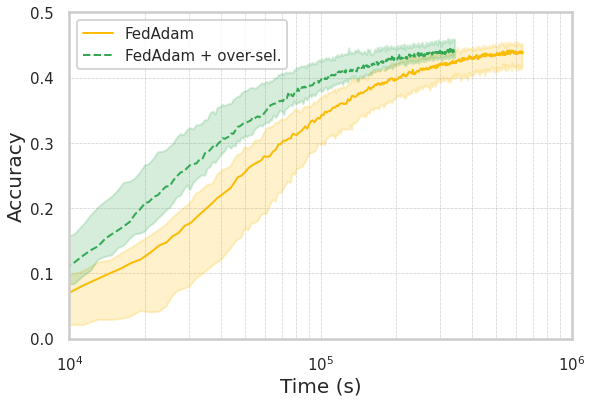}} \label{fig:fedadam_cifar_oversel_comparison_b}}%
  \\
  \subfloat[\centering \pdpe{} client latency]{{\includegraphics[width=0.48\linewidth]{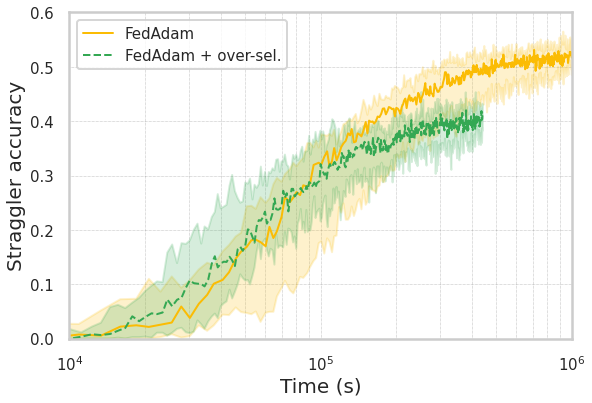}} \label{fig:fedadam_cifar_oversel_comparison_c}}%
  \subfloat[\centering \pdpe{} client latency]{{\includegraphics[width=0.48\linewidth]{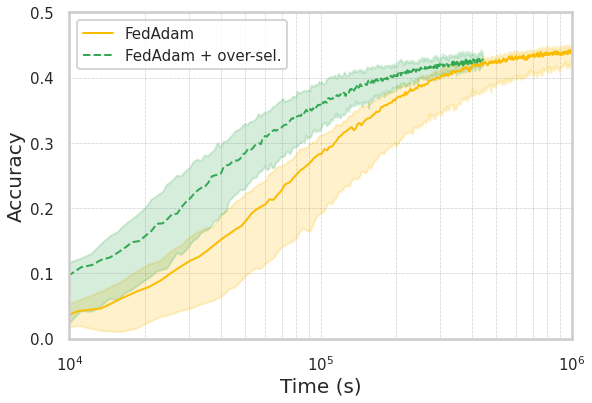}} \label{fig:fedadam_cifar_oversel_comparison_d}}%
  \caption{Straggler accuracy (left) and total accuracy (right) as a function wall clock training time for CIFAR-100 with the \pe{} client latency (top) and \pdpe{} client latency scenario (bottom). The training curves for the \fedadam{} algorithm are shown with and without over-selection.}%
  \label{fig:fedadam_cifar_oversel_comparison}%
\end{figure}

\paragraph{Impact of over-selection} Over-selection significantly sped up \fedavg{} and \fedadam{}, and even boosted accuracy on the \pe{} client latency tasks (Figure~\ref{fig:fedadam_cifar_oversel_comparison_a} and~\ref{fig:fedadam_cifar_oversel_comparison_b}). Over-selection was incredibly useful when there were no data domain differences between fast and straggler clients (only differences in data volume or other random factors). However, over-selection typically led to worse straggler accuracy and total accuracy on the \pdpe{} client latency task (Figure~\ref{fig:fedadam_cifar_oversel_comparison_c} and~\ref{fig:fedadam_cifar_oversel_comparison_d}). While fast and slow clients had the same data distribution in the \pe{} scenario, that was not the case for the \pdpe{} client latency scenario. In that setting, all of the clients with straggler data domains (0-4 for EMNIST, ``vehicles 1" and ``vehicles 2" for CIFAR-100, ``math" and ``c + +" for StackOverflow) were dropped from the training round due to high latency. In tuning experiments, we observed that the best trade-offs were achieved when the over-selection percentage corresponded roughly to the prevalence of straggler clients in the total population. 

\paragraph{Time-limited client computation} Limiting the client computations by time rather than the number of local epochs or local steps of SGD significantly sped up training for \fedavg{}. In the \pe{} client latency scenario, the speed-up often came with no reduction in accuracy (Table~\ref{table:emnist_pe_full}). However, in the \pdpe{} latency scenario, the speed-up came at the price of reduced straggler accuracy (Table~\ref{table:emnist_pdpe_full}). The difference was likely due to the fact that, in the \pdpe{} latency scenario, fewer training steps were performed on slow clients. This inevitably biased the learned models towards the fast clients that lacked straggler data domains. Furthermore, the trade-off between latency and accuracy was not competitive with other modifications to FL algorithms such as over-selection. Irreducible latency factors such as the communication time and system overhead were always present, and could be fully accounted for in the time limit applied by clients. While experiments were only performed for the \fedavg{} algorithm, the technique could be coupled with any of the other synchronous or asynchronous algorithms discussed in this work.

\paragraph{Accuracy vs. training time trade-off} Each algorithm has multiple parameters that adjust the trade-off between straggler accuracy, total accuracy, and total training time. To start with, the over-selection fraction for \fedavg{}, \fedadam{}, \faredust{}, and \feast{} could be increased to reduce training time, at the expense of straggler accuracy in the \pdpe{} latency scenario. For \fedbuff{}, the buffer size ($B$) and maximum number of concurrent clients ($\chi$) could be tuned to affect the accuracy-time trade-off. Larger $B$ typically led to smoother convergence and higher accuracy, but slower training. On the other hand, larger values of $\chi$ produced faster training but resulted in extremely high variance in model accuracy. Client computation time limits could also be used in conjunction with any algorithm to adjust the accuracy-time trade-off. However, experiments showed that other techniques such as over-selection provided better trade-offs in most tasks. Finally, the maximum wait time, $\tau_{\max}$, for the \feast{} algorithm affected the training time and straggler accuracy. $\tau_{\max}$ simply added a constant factor to the total training time. Even when waiting for 99\% of clients from all rounds, this amounted to a negligible increase in total training time. We observed that straggler accuracy increased with increasing $\tau_{\max}$.

\paragraph{StackOverflow experiments} \fedadam{} both with and without over-selection performed surprisingly well in experiments with the StackOverflow dataset. \fedadam{} achieved the highest straggler accuracy for StackOverflow with both client latency models (Table~\ref{table:so_pe_full} and~\ref{table:so_pdpe_full}), though it required an order of magnitude longer training time than the algorithms proposed in this work. \fedadam{} with over-selection, meanwhile, had comparable training time to \feast{} and \faredust{} with a straggler accuracy that was only modestly lower. We believe that the strong performance of the baseline algorithms on StackOverflow is due to the similarity of data domains for straggler clients and standard clients. Recall that straggler clients corresponded to users who submitted questions and answers with the ``math'' and ``c + +'' tags, while the standard clients corresponded to all other clients. Despite this tag separation, the language modeling tasks for the two groups of users were quite similar. There would be a high degree of overlap between the semantics and vocabularies of the two user domains, and no explicit output class separations. Thus, it is reasonable to expect models trained on standard clients in this setting could also generalize to straggler clients. This is in contrast to the EMNIST and CIFAR-100 datasets, where standard and straggler clients featured explicit output class partitions.
\section{Measurements from real-world FL} \label{sec:appendix_obs_latency}

Several real-world observations of FL client latency have been published previously. For example, \citet{wang2021field} estimated values for the parameters described in Section~\ref{sec:latency_model}. That work quoted average values in some cases and 90th percentile values for communication costs. Straggler modeling was highlighted as a direction for future work.

Client execution time histograms were presented in~\citet{huba2022papaya}. Execution times spanned 3 orders of magnitude in the examples provided. The mean round duration was also $21\times$ larger than the mean client execution time. Two factors were conflated in this result: the number of examples per device, and the per-example training time. Histograms of the number of examples per client also spanned 3 orders of magnitude. This suggested that variance in the number of training steps explained the latency distribution.

We measured client latencies in a production keyword spotting task similar to those presented in~\citet{leroy2019federated} and~\citet{hard2022production}. Percentile values for the client upload time, client train time, and the system constant time are shown in Table~\ref{table:real_world_latency}. Measurements were taken during peak training hours for a model consisting of approximately 300,000 parameters with a communication payload of 1.3MB. Approximately 600 examples were processed by the median client. As with the measurements from~\citet{huba2022papaya}, the train time quoted here conflates the number of steps per client and the per-step execution times. However, based on the number of client steps for the tasks, it was possible to infer that per-step computation times varied by $5-10\times$ between the median and 99th percentile.

\begin{table}
  \caption{The client upload time,  client train time, and system constant time as measured in a real-world application of FL for keyword spotting.}
  \label{table:real_world_latency}
  \vskip 0.15in
  \centering
  \begin{tabular}{crrc}
    \toprule
    Percentile & Upload [s] & Train [s] & System [s]\\
    \midrule 
    50\%       &  5.1         &  26        & 24 \\
    75\%       &  7.6         &  86        & 33 \\
    95\%       &  9.6         & 180        & 36 \\
    99\%       &   15         & 270        & 40 \\
    \bottomrule
  \end{tabular}
\end{table}

Latency component distributions vary with the model size, the federated computations, the client device types, and even the time of day. Nevertheless, some general observations can be made from the real-world examples. First, total training time for clients can vary by 1-2 orders of magnitude across a federated population. And second, all sources of client latency have large differences between the median and 99th percentile values. In the examples shown here, training time was a more dominant factor than communication time. But the trade-off would change for different model sizes or numbers of local client training steps.

\section{Limitations} \label{sec:appendix_limitations}

In this work, we compared the capacity of multiple FL algorithms to learn from straggler clients. The following section attempts to elucidate some of the assumptions and simplifications that went into the simulation framework (Appendix~\ref{sec:appendix_latency_model_limitations}) and algorithm comparisons (Appendix~\ref{sec:appendix_algo_comp_limitations}).

\subsection{Latency simulation limitations}
\label{sec:appendix_latency_model_limitations}

The Monte Carlo client latency simulations presented in Section~\ref{sec:latency} provide new tools for analyzing the role of fast and slow clients in simulations. That being said, there are limitations to the MC.

\begin{enumerate}
    \item The log-normal distribution worked well from a practical standpoint for parameterizing the latency factors. However, the individual latency factor distributions are highly dependent on the devices and characteristics of the clients involved in FL. This work would benefit from explorations of additional functional forms for the latency. Future works might also consider adopting distributions from production applications directly.
    \item Communication time was modeled as a single parameter ($T_{\textrm{Communication}}$) in this work. But it would be more realistic to decompose the communication time into upload and download times that scale with the bytes transferred in each direction. This would enable explorations of the trade-offs between client computation time and communication cost. In particular, this constant parameter affects latency comparisons with \faredust{}, which has a larger download cost as a result of the teacher model. It was determined that this additional latency term would have only a small effect on the overall simulated latency (on the order of a few percent) and would not change the observations presented in this work.
    \item The per-example computation cost was the same for all examples on a given client in this work. But for sequential modeling tasks (i.e. StackOverflow), the per-example compute time should scale with the number of sequential elements (such as tokens or time frames). This change would help explore scenarios in which straggler clients are defined by the length of data sequences, rather than just the number or type of sequences as in this work.
    \item This work assumes that the communication latency and per-example training latency distributions are uncorrelated. However, real-world applications of FL probably have more complex correlations. New devices, for instance, may process training examples more rapidly and typically support faster network connections.
    \item The role of client sampling was not explored in this work. In practice, clients in FL are not sampled uniformly at random from the population. In fact, it is quite likely that straggler clients are also less likely to check-in for training, presenting yet another obstacle to learning from them in production settings. Explorations of the relationship between client sampling probability and client latency are possible directions for future work.
\end{enumerate}

\subsection{Algorithm comparison limitations}
\label{sec:appendix_algo_comp_limitations}

We endeavored to provide fair comparisons of existing FL algorithms (\fedavg{}, \fedadam{}, and \fedbuff{}) with our new algorithms (\faredust{} and \feast{}) on a variety of benchmark tasks. Some limitations of our tuning procedures and comparisons of results are summarized below.

\begin{enumerate}
    \item Results with time-limited client computations were only reported for one time limit (the 75th percentile client training time for one local epoch). We did not scan the value of the client computation time limit, which would have yielded a range of accuracy-time trade-offs.
    \item Upload and download times were excluded from the client computation time limits, meaning that stragglers with high communication latencies were not helped with this technique. While incorporating (post-training) upload times is impossible, it might be possible to estimate future upload times and use those to set time limits on the overall client computation process.
    \item All of the algorithms presented have many levers to tune the accuracy-time trade-offs. Examples include the over-selection fraction, the max wait time for \feast{}, the buffer size and max concurrency for \fedbuff{}, and more. We did not present complete scans of these values, which would have elucidated the accuracy-time trade-offs. Instead, we identified individual working points which we believed to be representative of the best trade-offs provided by each algorithm.
    \item The \fedbuff{} algorithm was tuned for each dataset and latency model. However, we failed to find a configuration for StackOverflow that converged smoothly and rapidly. We believe that increasing the max concurrency beyond the values which we scanned, combined with larger buffer sizes, might lead to faster convergence than in the results that were reported.
    \item Similarly, \feast{} was tuned for each dataset and latency model. In several cases, the model was over-tuned for straggler accuracy. That is, it achieved remarkably high straggler accuracy, but at the cost of significantly lower total accuracy. We believe that better trade-offs between total accuracy and straggler accuracy could be identified with a modified hyper-parameter tuning objective.
    \item For hyper-parameter selection, we chose to maximize the straggler accuracy. While this ultimately produced models with higher overall accuracy in many cases, it would be better to identify a more balanced hyper-parameter tuning objective in future works.
\end{enumerate}

\section{Broader impact} \label{sec:appendix_impact}






The FL literature has noted (e.g. in \citet{wang2021field, kairouz2021advances}) that while the FL platform enables the possibility of broader inclusion of data from minority populations, it is generally a non-trivial matter to correct for biases arising from systems and data heterogeneity. \citet{kairouz2021advances} states that `` ... output from devices with faster processors may be over-represented, with these devices likely newer devices and thus correlated with socioeconomic status''. One of the most prevalent forms of FL is one where the participating client devices are mobile phones, and correlations between mobile phone usage and socioeconomic status have been identified (\citet{frias2012relationship}, \citet{steele2021mobility}).

Against this backdrop, the improvements proposed in this paper to better include straggler clients are not merely motivated by improved learning efficiency, but by the potential benefits of reducing bias in FL model training and evaluation. A beneficial societal impact of the production deployment of this work would be improved inclusion of data from poorer populations (via improved inclusion of their `straggling' older model mobile phones) and less-connected populations (via improved inclusion of their `straggling' slower internet connections). 

In general, a tension exists in ML and data science between privacy and fairness (e.g., see Section 7.4 in \citet{wang2021field}). However, a nice property of the methods described in this paper is that they address data inclusion via a simple and non-private discriminating metric: a client's total round run time $T_{\text{round}}$. We only focus on methods which allow for improved contribution from clients with the largest $T_{\text{round}}$ in a given cohort. There's no need to do potentially privacy-impacting actions such as e.g. identifying and clustering clients by their data similarity.

\section{Simulated client latency distributions} \label{sec:appendix_latency_dist}

This section provides additional information about the simulated client latency distributions introduced in Section~\ref{sec:latency} as they were applied to the EMNIST, CIFAR-100, and StackOverflow tasks described in Section~\ref{sec:fl_tasks}. A summary of the client statistics for each task is also provided in Table~\ref{table:task_client_statistics}.

\begin{table}
  \caption{Client statistics for the EMNIST, CIFAR-100, and StackOverflow datasets as described in Section~\ref{sec:data_and_tasks}.}
  \label{table:task_client_statistics}
  \vskip 0.15in
  \centering
  \begin{tabular}{lccc}
    \toprule
    Task & Standard clients & Straggler clients & Total clients \\
    \midrule 
    EMNIST        &   2,600 &    800 &   3,400 \\
    CIFAR-100     &   375   &    125 &     500 \\
    StackOverflow & 248,595 & 93,882 & 342,477 \\
    \bottomrule
  \end{tabular}
\end{table}

\subsection{EMNIST client latency distributions}

\begin{figure}%
  \centering
  \subfloat[\centering \pe{} client latency]{{\includegraphics[width=0.45\linewidth]{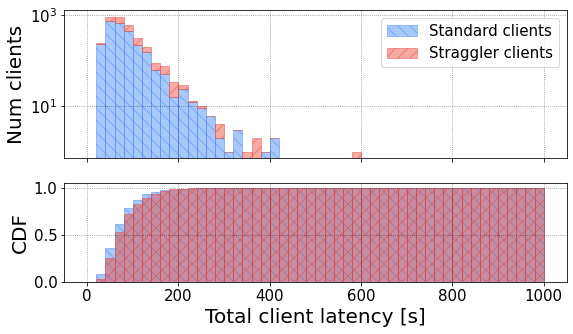} \label{fig:emnist_dist_appendix_a}}}%
  \qquad
  \subfloat[\centering \pdpe{} client latency]{{\includegraphics[width=0.45\linewidth]{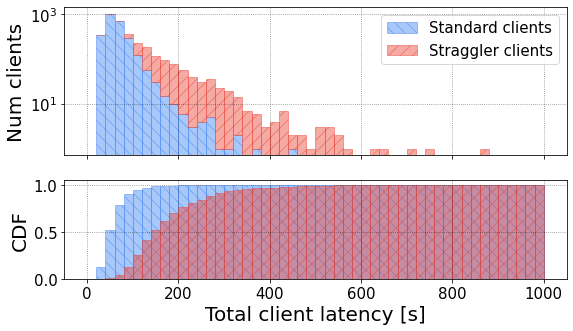} \label{fig:emnist_dist_appendix_b}}}%
  \caption{The distribution of total client latencies, in seconds, for the EMNIST dataset. The PDFs (top) are stacked, while the CDFs (bottom) are overlaid. In the \pe{} client latency model (Figure~\ref{fig:emnist_dist_appendix_a}), client latency is a function of the number of training examples per client only, and not data domain. As a result, the standard client distribution and straggler client distribution closely match. In the \pdpe{} client latency model (Figure~\ref{fig:emnist_dist_appendix_b}), latency is a function of the number of training examples per client as well as the data domains of the clients. As a result, the straggler client distributions have significantly higher mean values and longer tails than the standard clients.}%
  \label{fig:emnist_dist_appendix}%
\end{figure}

EMNIST client latency distributions are shown in Figure~\ref{fig:emnist_dist_appendix}, and summary statistics are provided in Table~\ref{table:emnist_percentiles_appendix}. Recall that for EMNIST, straggler clients were defined as the 800 clients with the most examples from digits 0-4, while standard clients were the remaining 2,600 clients that have examples with digits 0-4 removed. For the \pe{} scenario, straggler clients and standard clients shared the same log-normal parameterizations of the latency factors. As a result, straggler clients had comparable latency distributions to standard clients. The median straggler client had slightly more examples than the median standard client (77.14 seconds vs. 69.89 seconds median latency, respectively), due to the fact that examples featuring the digits 0-4 were removed from standard client data partitions. This explains the slightly higher median latency in this scenario. However, both sets of clients had 99th percentile latencies close to 223 seconds.

For the EMNIST \pdpe{} client latency scenario, the straggler clients had log-normal parameterizations of the latency factors with larger values of $\mu$ and $\sigma$. Consequently, the straggler client latencies were typically much longer than those of standard clients. The median straggler client latency was nearly $3\times$ greater than the median standard client latency (151.5 seconds vs 59.35 seconds, respectively).

\begin{table}
  \caption{Simulated client latencies with the EMNIST dataset, using the \pe{} client latency model (left) and \pdpe{} client latency model (right).}
  \label{table:emnist_percentiles_appendix}
  \vskip 0.15in
  \centering
  \begin{tabular}{lcccccc}
    \toprule
    & \multicolumn{3}{c}{\pe{}} & \multicolumn{3}{c}{\pdpe{}} \\
    \cmidrule(r){2-4} \cmidrule(r){5-7}
    & 50\% [s] & 95\% [s] & 99\% [s] & 50\% [s] & 95\% [s] & 99\% [s] \\
    \midrule 
    Standard clients  & 69.89 & 151.7 & 223.4 & 59.35 & 117.3 & 194.3 \\
    Straggler clients & 77.14 & 160.5 & 223.8 & 151.5 & 323.8 & 472.0 \\
    \bottomrule
  \end{tabular}
\end{table}

\subsection{CIFAR-100 client latency distributions}

\begin{figure}%
  \centering
  \subfloat[\centering \pe{} client latency]{{\includegraphics[width=0.45\linewidth]{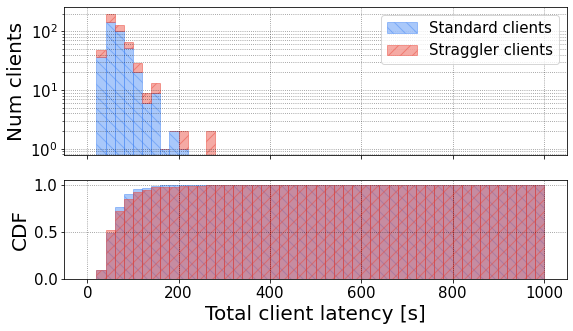} \label{fig:cifar_dist_appendix_a}}}%
  \qquad
  \subfloat[\centering \pdpe{} client latency]{{\includegraphics[width=0.45\linewidth]{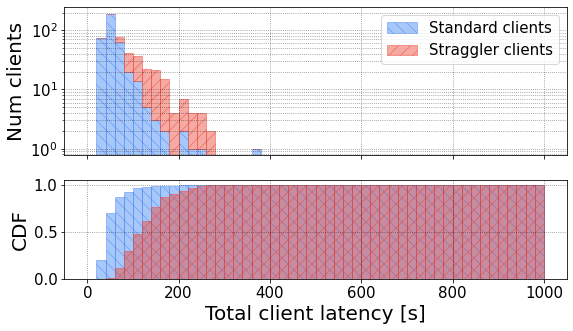} \label{fig:cifar_dist_appendix_b}}}%
  \caption{The distribution of total client latencies, in seconds, for the CIFAR-100 dataset. The PDFs (top) are stacked, while the CDFs (bottom) are overlaid. In the \pe{} client latency model (Figure~\ref{fig:cifar_dist_appendix_a}), client latency is a function of the number of training examples per client only, and not data domain. As a result, the standard client distribution and straggler client distribution closely match. In the \pdpe{} client latency model (Figure~\ref{fig:cifar_dist_appendix_b}), latency is a function of the number of training examples per client as well as the data domains of the clients. As a result, the straggler client distributions have significantly higher mean values and longer tails than the standard clients.}%
  \label{fig:cifar_dist_appendix}%
\end{figure}

CIFAR-100 client latency distributions are shown in Figure~\ref{fig:cifar_dist_appendix}, and summary statistics are provided in Table~\ref{table:cifar_percentiles_appendix}. Recall that for CIFAR-100, straggler clients were defined as the 125 clients with the most examples from the ``vehicles 1'' and ``vehicles 2'' coarse classes, while standard clients were the remaining 375 clients that had examples with those classes removed. For the \pe{} scenario, straggler clients and standard clients shared the same log-normal parameterizations of the latency factors. As a result, straggler clients had nearly the same latency distributions as standard clients across all latency percentiles.

For the CIFAR-100 \pdpe{} client latency scenario, the straggler clients had log-normal parameterizations of the latency factors with larger values of $\mu$ and $\sigma$. Consequently, the straggler client latencies were typically much longer than those of standard clients. The median straggler client latency (114.2 seconds) was slower than the 95th percentile standard client latency (111.2 seconds). And stragglers were $3\times$ slower than standard clients at the 99th percentile (450.6 seconds vs. 154.6 seconds, respectively).

\begin{table}
  \caption{Simulated client latencies with the CIFAR-100 dataset, using the \pe{} client latency model (left) and \pdpe{} client latency model (right).}
  \label{table:cifar_percentiles_appendix}
  \vskip 0.15in
  \centering
  \begin{tabular}{lcccccc}
    \toprule
    & \multicolumn{3}{c}{\pe{}} & \multicolumn{3}{c}{\pdpe{}} \\
    \cmidrule(r){2-4} \cmidrule(r){5-7}
    & 50\% [s] & 95\% [s] & 99\% [s] & 50\% [s] & 95\% [s] & 99\% [s] \\
    \midrule
    Standard clients  & 60.21 & 117.9 & 193.5 & 52.04 & 111.2 & 154.6 \\
    Straggler clients & 60.21 & 117.1 & 204.7 & 114.2 & 259.9 & 450.6 \\
    \bottomrule
  \end{tabular}
\end{table}

\subsection{StackOverflow client latency distributions}

\begin{figure}%
  \centering
  \subfloat[\centering \pe{} client latency]{{\includegraphics[width=0.45\linewidth]{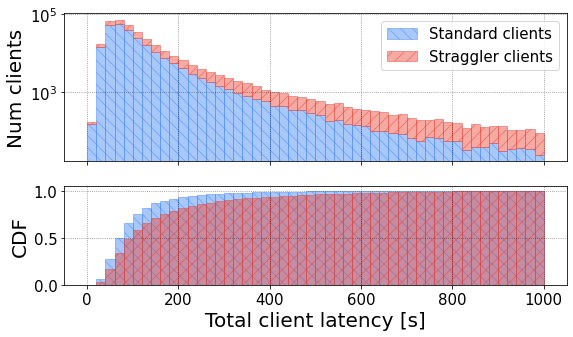} \label{fig:stackoverflow_dist_appendix_a}}}%
  \qquad
  \subfloat[\centering \pdpe{} client latency]{{\includegraphics[width=0.45\linewidth]{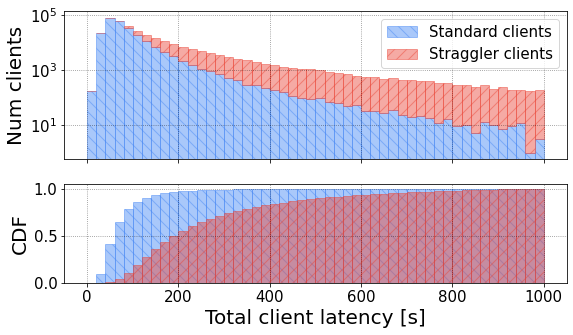} \label{fig:stackoverflow_dist_appendix_b}}}%
  \caption{The distribution of total client latencies, in seconds, for the StackOverflow dataset. The PDFs (top) are stacked, while the CDFs (bottom) are overlaid. In the \pe{} client latency model (Figure~\ref{fig:stackoverflow_dist_appendix_a}), client latency is a function of the number of training examples per client only, and not data domain. As a result, the standard domain client distribution and straggler domain client distribution are comparable. The straggler client distribution is a bit longer tailed because clients in the dataset containing questions or answers with the ``math'' or ``c + +'' tags had a greater number of associated examples. In the \pdpe{} client latency model (Figure~\ref{fig:stackoverflow_dist_appendix_b}), latency is a function of the number of training examples per client as well as the data domains of the clients. As a result, the straggler domain client distributions have significantly higher mean values and longer tails than the standard domain clients.}%
  \label{fig:stackoverflow_dist_appendix}%
\end{figure}

StackOverflow client latency distributions are shown in Figure~\ref{fig:stackoverflow_dist_appendix}, and summary statistics are provided in Table~\ref{table:so_percentiles_appendix}. Recall that for StackOverflow, straggler clients were defined as the 93,882 clients with questions or answers featuring the ``math'' or ``c + +'' tags, while the remaining 248,595 standard clients did not feature such examples. For the \pe{} scenario, straggler clients and standard clients shared the same log-normal parameterizations of the latency factors. Even so, straggler clients had longer tailed latency distributions than standard clients. This was due to the fact that authors of ``math'' and ``c + +'' posts submitted more questions and answers than the average author, resulting in larger simulated client datasets for stragglers than standard clients. The median straggler client had a total latency of 106.2 seconds, while the median standard client had a latency of 80.39 seconds. But the 99th percentile straggler client latency was 1,811 seconds, significantly greater than the 520.1 seconds for standard clients at the 99th percentile.

For the StackOverflow \pdpe{} client latency scenario, the straggler clients had log-normal parameterizations of the latency factors with larger values of $\mu$ and $\sigma$. Consequently, the straggler client latencies were even longer than those of standard clients. The median straggler client latency (211.2 seconds) was slower than the 95th percentile standard client latency (181.7 seconds). And stragglers were $10\times$ slower than standard clients at the 99th percentile (3,369 seconds vs. 325.9 seconds, respectively).

\begin{table}
  \caption{Simulated client latencies with the StackOverflow dataset, using the \pe{} client latency model (left) and \pdpe{} client latency model (right).}
  \label{table:so_percentiles_appendix}
  \vskip 0.15in
  \centering
  \begin{tabular}{lcccccc}
    \toprule
    & \multicolumn{3}{c}{\pe{}} & \multicolumn{3}{c}{\pdpe{}} \\
    \cmidrule(r){2-4} \cmidrule(r){5-7}
    & 50\% [s] & 95\% [s] & 99\% [s] & 50\% [s] & 95\% [s] & 99\% [s] \\
    \midrule
    Standard clients  & 80.39 & 255.8 & 520.1 & 66.20 & 181.7 & 325.9 \\
    Straggler clients & 106.2 & 615.5 & 1,811 & 211.2 & 1,167 & 3,369 \\
    \bottomrule
  \end{tabular}
\end{table}

\section{Algorithm tuning} \label{sec:tuning}

The \fedbuff{}, \faredust{}, and \feast{} algorithms were tuned individually for each task (EMNIST, CIFAR-100, and StackOverflow) and latency model (\pe{} and \pdpe{}). This section describes the hyper-parameter values that were scanned during tuning for each algorithm in order to maximize the straggler accuracy. We also discuss the \fedavg{} and \fedadam{} hyper-parameters that were adopted from~\citet{reddi2020adaptive}.

\subsection{Synchronous \fedavg{} and \fedadam{} baselines} \label{sec:server_optimizer_baselines}

Experiments with \fedavg{} and \fedadam{} primarily utilized the optimal hyper-parameters identified in~\citet{reddi2020adaptive} for the EMNIST, CIFAR-100, and StackOverflow tasks. Two parameters --- cohort size and the over-selection cohort size --- were modified in order to provide more consistency between experiment trials. Table~\ref{table:fedavg_defaults_from_afo} provides the hyper-parameter values for \fedavg{}, while Table~\ref{table:fedadam_defaults_from_afo} provides the hyper-parameter values for \fedadam{}. Note that the same hyper-parameter values were used for each task in experiments with both the \pe{} and \pdpe{} client latency scenarios.

\begin{table}
  \caption{The default hyper-parameters for \fedavg{} in this work were based on those quoted in~\citet{reddi2020adaptive}. Only the cohort size and over-selection cohort size were modified. The parameters below were used for experiments with both the \pe{} and \pdpe{} latency scenarios for each task.}
  \label{table:fedavg_defaults_from_afo}
  \vskip 0.15in
  \centering
  \begin{tabular}{lccc}
    \toprule
    Parameter & EMNIST & CIFAR-100 & StackOverflow \\
    \midrule 
    Client batch size                  &     20 &  20 &      16 \\
    Client number of epochs            &      1 &   1 &       1 \\
    Client learning rate ($\eta_l$)    &    0.1 & 0.1 &     0.3 \\
    Server learning rate ($\eta_g$)    &    1.0 & 3.0 &     1.0 \\
    Cohort size ($B$)                  &     50 &  20 &     100 \\
    Over-selection cohort size ($B_t$) &     60 &  24 &     120 \\
    \bottomrule
  \end{tabular}
\end{table}

\begin{table}
  \caption{The default hyper-parameters for \fedadam{} in this work were also based on those quoted in~\citet{reddi2020adaptive}. Only the cohort size and over-selection cohort size were adjusted. The parameters below were used for experiments with both the \pe{} and \pdpe{} latency scenarios for each task.}
  \label{table:fedadam_defaults_from_afo}
  \vskip 0.15in
  \centering
  \begin{tabular}{lccc}
    \toprule
    Parameter & EMNIST & CIFAR-100 & StackOverflow \\
    \midrule
    Client batch size                  &     20 &   20 &      16 \\
    Client number of epochs            &      1 &    1 &       1 \\
    Client learning rate ($\eta_l$)    &   0.03 & 0.03 &     0.3 \\
    Server learning rate ($\eta_g$)    &  0.003 &  1.0 &    0.01 \\
    Server Adam $\beta_{1}$            &    0.9 &  0.9 &     0.9 \\
    Server Adam $\beta_{2}$            &   0.99 & 0.99 &    0.99 \\
    Server Adam $\epsilon$             & $10^{-4}$ &  $10^{-1}$ & $10^{-5}$ \\
    Cohort size ($B$)                  &     50 &   20 &     100 \\
    Over-selection cohort size ($B_t$) &     60 &   24 &     120 \\
    \bottomrule
  \end{tabular}
\end{table}

\subsection{Asynchronous \fedbuff{} baseline tuning}

The \fedbuff{} algorithm~\cite{nguyen2022federated} as described in Section~\ref{sec:algorithms} was optimized using two sequential 2D grid searches: first a 2D search of the client learning rate, $\eta_l$, and server learning rate, $\eta_g$, followed by a search of the number of clients per server update (a.k.a. the buffer size, $B$), and the maximum client concurrency, $\chi$. The tuned values of the parameters for \fedbuff{} along with the specific operating points considered during tuning are provided in Table~\ref{table:fedbuff_tuning_pe} for the \pe{} client latency scenario and Table~\ref{table:fedbuff_tuning_pdpe} for the \pdpe{} client latency scenario.

\begin{table}
  \caption{The tuned hyper-parameter values for \fedbuff{} with the \pe{} latency model, along with the operating points that were considered for each parameter. Tuning was performed independently for each task.}
  \label{table:fedbuff_tuning_pe}
  \vskip 0.15in
  \centering
  \begin{tabular}{lcccc}
    \toprule
    & & \multicolumn{3}{c}{Tuned value} \\
    \cmidrule(r){3-5}
    Parameter & Values scanned & EMNIST & CIFAR-100 & StackOverflow \\
    \midrule 
    $\eta_l$ & $\{ 10^{-4}, 10^{-3}, 10^{-2}, 10^{-1}, 0.3, 1 \}$ & $10^{-1}$ & $10^{-1}$ & 0.3 \\
    $\eta_g$ & $\{ 10^{-4}, 10^{-3}, 10^{-2}, 10^{-1}, 0.3, 1 \}$ & 1 & 1 & 1 \\
    $B$    & $\{ 5, 10, 20, 50 \}$ &  20 & 20 &  50 \\
    $\chi$ & $\{ 60, 100, 200 \}$  & 100 & 60 & 100 \\
    \bottomrule
  \end{tabular}
\end{table}

\begin{table}
  \caption{The tuned hyper-parameter values for \fedbuff{} with the \pdpe{} latency model, along with the operating points that were considered for each parameter. Tuning was performed independently for each task.}
  \label{table:fedbuff_tuning_pdpe}
  \vskip 0.15in
  \centering
  \begin{tabular}{lcccc}
    \toprule
    & & \multicolumn{3}{c}{Tuned value} \\
    \cmidrule(r){3-5}
    Parameter & Values scanned & EMNIST & CIFAR-100 & StackOverflow \\
    \midrule 
    $\eta_l$ & $\{ 10^{-4}, 10^{-3}, 10^{-2}, 10^{-1}, 1 \}$ & $10^{-2}$ & $10^{-1}$ & 1 \\
    $\eta_g$ & $\{ 10^{-4}, 10^{-3}, 10^{-2}, 10^{-1}, 1 \}$ &  1 & 1 & $10^{-1}$ \\
    $B$    & $\{ 5, 10, 20, 50 \}$ &   20 & 20 &  50 \\
    $\chi$ & $\{ 60, 100, 200 \}$  &  200 & 60 & 100 \\
    \bottomrule
  \end{tabular}
\end{table}

\fedbuff{} was also explored in conjunction with distillation regularization (with strength parameter $\rho$), proximal $L_2$ regularization (with strength parameter $\nu$, and EMA post-processing of weights (with decay strength $\beta$). A 1-dimensional scan was performed to tune each parameter once $\eta_g$, $\eta_l$, $B$, and $\chi$ had already been tuned and fixed. The scanned operating points, as well as the tuned values, are shown in Table~\ref{table:fedbuff_addons_pe} for the \pe{} client latency scenario and Tables~\ref{table:fedbuff_addons_pdpe} for the \pdpe{} client latency scenario.

\begin{table}
  \caption{The tuned hyper-parameter values for \fedbuff{} with the \pe{} client latency model, along with the operating points that were considered. $\beta$ is the EMA decay, $\rho$ is the knowledge distillation regularization strength, and $\nu$ is the proximal $L_2$ regularization strength. Tuning was performed independently for each task.}
  \label{table:fedbuff_addons_pe}
  \vskip 0.15in
  \centering
  \begin{tabular}{lcccc}
    \toprule
    & & \multicolumn{3}{c}{Tuned value} \\
    \cmidrule(r){3-5}
    Parameter & Values scanned & EMNIST & CIFAR-100 & StackOverflow \\
    \midrule 
    $\beta$  & $\{ 0, 0.9, 0.99, 0.999 \}$              & 0.99 & 0.9 & 0.99 \\
    $\rho$ & $\{ 0, 10^{-4}, 10^{-3}, 10^{-2}, 0.1, 0.3, 0.5 \}$ &  0.1 & 0.1 & $10^{-2}$  \\
    $\nu$    & $\{ 0, 10^{-4}, 10^{-3}, 10^{-2}, 0.1, 0.3, 0.5 \}$ &  0.5 & 0.3 & $10^{-2}$  \\
    \bottomrule
  \end{tabular}
\end{table}

\begin{table}
  \caption{The tuned hyper-parameter values for \fedbuff{} with the \pdpe{} client latency model, along with the operating points that were considered. $\beta$ is the EMA decay, $\rho$ is the knowledge distillation regularization strength, and $\nu$ is the proximal $L_2$ regularization strength. Tuning was performed independently for each task.}
  \label{table:fedbuff_addons_pdpe}
  \vskip 0.15in
  \centering
  \begin{tabular}{lcccc}
    \toprule
    & & \multicolumn{3}{c}{Tuned value} \\
    \cmidrule(r){3-5}
    Parameter & Values scanned & EMNIST & CIFAR-100 & StackOverflow \\
    \midrule 
    $\beta$  & $\{ 0, 0.9, 0.99, 0.999 \}$         & 0.99 & 0.9 & 0.99 \\
    $\rho$ & $\{ 0, 10^{-4}, 10^{-3}, 10^{-2}, 0.05, 0.1, 0.3 \}$ & 0.05 & 0.1 & $10^{-2}$  \\
    $\nu$    & $\{ 0, 10^{-4}, 10^{-3}, 10^{-2}, 0.05, 0.1, 0.3 \}$ & 0.05 & 0.3 & $10^{-4}$  \\
    \bottomrule
  \end{tabular}
\end{table}

\subsection{\faredust{} tuning}

Three parameters were tuned for the \faredust{} algorithm introduced in Section~\ref{sec:fare_dust}: the number of historical teachers ($k$), the EMA decay strength ($\beta$), and the distillation regularization strength ($\rho$). The tuning search was conducted on the outer product of the operating points listed in Table~\ref{table:fare_dust_params_pe}. The optimal values are also provided in Table~\ref{table:fare_dust_params_pe} for the \pe{} client latency scenario and Table~\ref{table:fare_dust_params_pdpe} for the \pdpe{} client latency scenario. \faredust{} was used in conjunction with the \fedadam{} server optimizer. Default values for \fedadam{} from Appendix~\ref{sec:server_optimizer_baselines} were used without tuning for all other hyper-parameters.

\begin{table}
  \caption{The operating points scanned for each \faredust{} parameter, as well as the optimal values identified for all tasks with the \pe{} latency model. $k$ represents the number of historical teachers, $\beta$ governs the EMA decay strength, and $\rho$ represents the distillation regularization strength.}
  \label{table:fare_dust_params_pe}
  \vskip 0.15in
  \centering
  \begin{tabular}{lcccc}
    \toprule
    & & \multicolumn{3}{c}{Tuned value} \\
    \cmidrule(r){3-5}
    Par.  & Values scanned & EMNIST & CIFAR-100 & StackOverflow \\
    \midrule 
    $k$     & $\{ 1, 2, 5, 10, 20, 50, 200, 500, 2000 \}$                      &   50 &  50 &    100 \\
    $\beta$ & $\{ 0.0, 0.9, 0.95, 0.99, 0.995, 0.999 \}$                       & 0.99 & 0.9 &   0.99 \\
    $\rho$  & $\{ 0, 10^{-4}, 5\times10^{-4}, 10^{-3}, 10^{-2}, 10^{-1}, 1 \}$ &  0.1 & 0.1 & 0.0005 \\
    \bottomrule
  \end{tabular}
\end{table}

\begin{table}
  \caption{The operating points scanned for each \faredust{} parameter, as well as the optimal values identified for all tasks with the \pdpe{} latency model. $k$ represents the number of historical teachers, $\beta$ governs the EMA decay strength, and $\rho$ represents the distillation regularization strength.}
  \label{table:fare_dust_params_pdpe}
  \vskip 0.15in
  \centering
  \begin{tabular}{lcccc}
    \toprule
    & & \multicolumn{3}{c}{Tuned value} \\
    \cmidrule(r){3-5}
    Par.  & Values scanned & EMNIST & CIFAR-100 & StackOverflow \\
    \midrule 
    $k$     & $\{ 1, 2, 5, 10, 20, 50, 200, 500, 2000 \}$                      &   50 &  50 &    100 \\
    $\beta$ & $\{ 0, 0.9, 0.95, 0.99, 0.995, 0.999 \}$                         & 0.99 & 0.9 &   0.99 \\
    $\rho$  & $\{ 0, 10^{-4}, 5\times10^{-4}, 10^{-3}, 10^{-2}, 10^{-1}, 1 \}$ &  0.1 & 0.1 & 0.0005 \\
    \bottomrule
  \end{tabular}
\end{table}

\subsection{\feast{} tuning}

Three parameters were tuned for the \feast{} algorithm introduced in Section~\ref{sec:feast_on_msg}: the server learning rate $\eta_g$, the EMA decay strength $\beta$, and the auxiliary learning rate $\eta_{a}$. Instead of directly tuning $\eta_a$, we formed $\eta_a = \kappa\,\eta_g$ and tuned $\kappa \in [0, 1]$. Tuning was performed on the outer product of the operating points listed in Table~\ref{table:feast_on_msg_params_pe}. The optimal values identified by this grid search are also provided in Table~\ref{table:feast_on_msg_params_pe} for the \pe{} client latency model and Table~\ref{table:feast_on_msg_params_pdpe} for the \pdpe{} client latency model. The maximum wait time was set to an extremely high (but finite) value of $\tau_{\text{max}}=100,000$ seconds for all experiments. The \feast{} algorithm was used in conjunction with the \fedavg{} server optimizer function. Other than the explicitly specified exceptions, the default hyper-parameter values for \fedavg{} from Appendix~\ref{sec:server_optimizer_baselines} were used for experiments with \feast{}.

\begin{table}
  \caption{The \feast{} algorithm parameters, listed with the operating points scanned during tuning and the optimal values for each task with the \pe{} client latency scenario. $\eta_{g}$ represents the server learning rate, $\beta$ is the EMA decay strength, and $\kappa\ = (\eta_a / \eta_g)$ governs the auxiliary learning rate.}
  \label{table:feast_on_msg_params_pe}
  \vskip 0.15in
  \centering
  \begin{tabular}{lcccc}
    \toprule
    & & \multicolumn{3}{c}{Tuned value} \\
    \cmidrule(r){3-5}
    Par.     & Values scanned & EMNIST & CIFAR-100 & StackOverflow \\
    \midrule 
    $\eta_{g}$ & $\{0.003,0.006,0.01,0.03,0.06,0.1,0.2\}$ & 0.006 &   0.2 &  0.01 \\
    $\beta$    & $\{0.9,0.95,0.99,0.995,0.999\}$          &  0.95 & 0.995 & 0.995 \\
    $\kappa$   & $\{0,0.2,0.5,0.8,0.9,1.0\}$              &     0 &   0.5 &     0 \\
    \bottomrule
  \end{tabular}
\end{table}

\begin{table}
  \caption{The \feast{} algorithm parameters, listed with the operating points scanned during tuning and the optimal values for each task with the \pdpe{} client latency scenario. $\eta_{g}$ represents the server learning rate, $\beta$ is the EMA decay strength, and $\kappa\ = (\eta_a / \eta_g)$ governs the auxiliary learning rate.}
  \label{table:feast_on_msg_params_pdpe}
  \vskip 0.15in
  \centering
  \begin{tabular}{lcccc}
    \toprule
    & & \multicolumn{3}{c}{Tuned value} \\
    \cmidrule(r){3-5}
    Par.     & Values scanned & EMNIST & CIFAR-100 & StackOverflow \\
    \midrule 
    $\eta_{g}$ & $\{0.003,0.006,0.01,0.03,0.06,0.1,0.2\}$ & 0.006 &   0.2 & 0.006 \\
    $\beta$    & $\{0.9,0.95,0.99,0.995,0.999\}$          &  0.99 & 0.995 &  0.95 \\
    $\kappa$   & $\{0,0.2,0.5,0.8,0.9,1.0\}$              &   0.9 &   0.5 &   0.2 \\
    \bottomrule
  \end{tabular}
\end{table}

\section{Full results} \label{sec:full_results}

This section contains the complete results for experiments with the EMNIST, CIFAR-100, and StackOverflow tasks combined with the \pe{} and \pdpe{} client latency scenarios. 10 independent trials were performed using identical hyper-parameters but different random initializations of the trainable variables for each algorithm, task, and latency scenario. Summaries of final evaluation metrics quote the median and 90\% confidence interval values from these 10 trials. 

\begin{figure}%
  \centering
  \subfloat[\centering Straggler accuracy as a function of wall clock training time.]{{\includegraphics[width=0.95\linewidth]{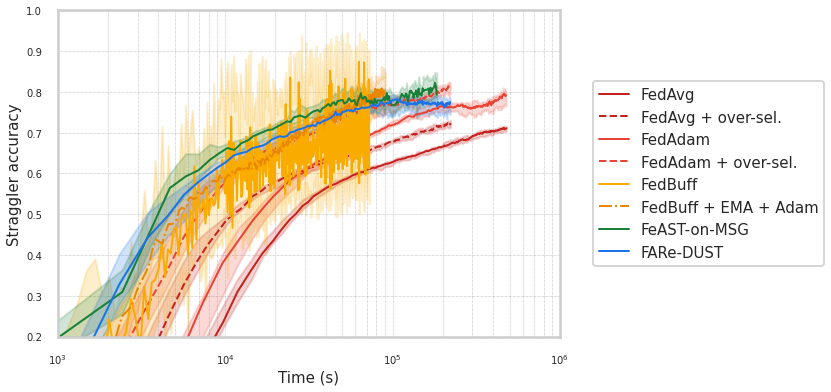}} \label{fig:acc_straggler_emnist_pe_all}}%
  \\
  \subfloat[\centering Total accuracy as a function of wall clock training time.]{{\includegraphics[width=0.95\linewidth]{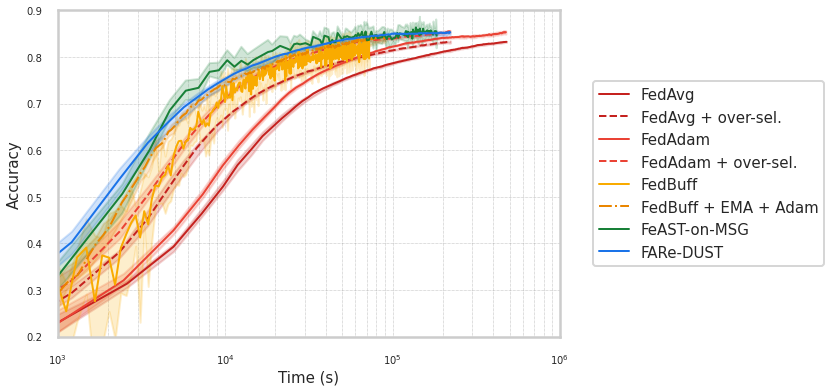}} \label{fig:acc_total_emnist_pe_all}}%
  \caption{Straggler accuracy (\ref{fig:acc_straggler_emnist_pe_all}) and total accuracy (\ref{fig:acc_total_emnist_pe_all}) as a function of wall clock training time for EMNIST with the \pe{} client latency scenario. Median and 90\% confidence interval values from 10 trials are illustrated with lines and bands, respectively.}%
  \label{fig:combined_acc_emnist_pe_all}%
\end{figure}

\begin{figure}%
  \centering
  \subfloat[\centering Straggler accuracy as a function of total accuracy.]{{\includegraphics[width=0.95\linewidth]{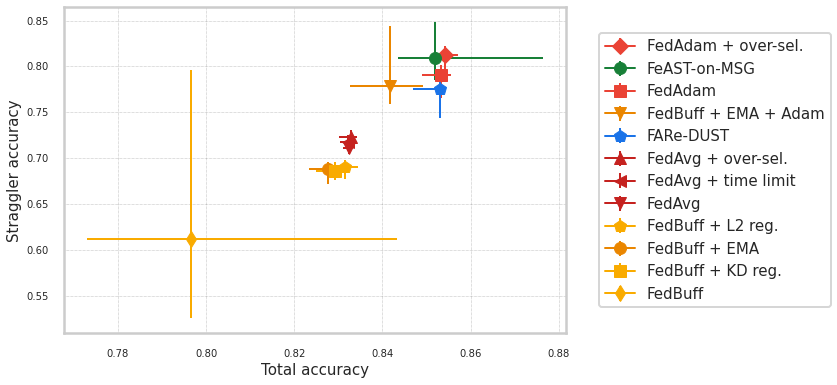}} \label{fig:2d_acc_emnist_pe_all}}%
  \\
  \subfloat[\centering Straggler accuracy as a function of total training time.]{{\includegraphics[width=0.95\linewidth]{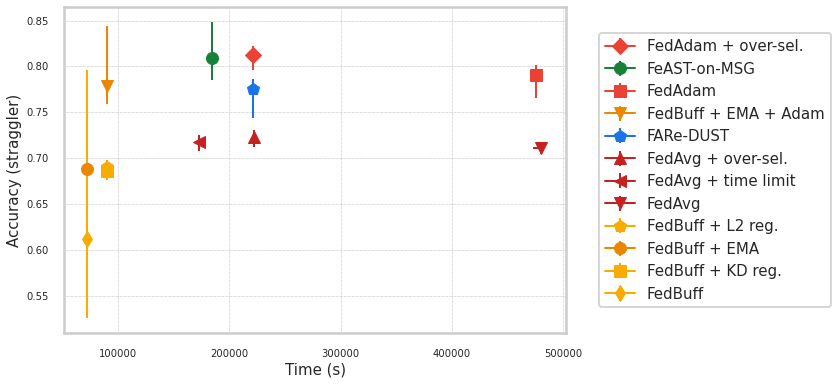}} \label{fig:2d_time_emnist_pe_all}}%
  \caption{Final straggler accuracy as a function of total accuracy (\ref{fig:2d_acc_emnist_pe_all}) and as a function of total wall clock training time (\ref{fig:2d_time_emnist_pe_all}) for EMNIST with the \pe{} client latency scenario. Median and 90\% confidence interval values from 10 trials are illustrated with points and error bars, respectively.}%
  \label{fig:combined_2d_emnist_pe_all}%
\end{figure}

\begin{figure}%
  \centering
  \subfloat[\centering Straggler accuracy as a function of wall clock training time.]{{\includegraphics[width=0.95\linewidth]{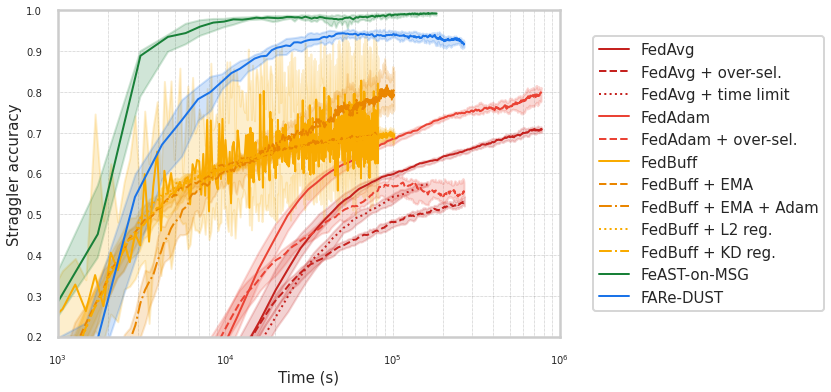}} \label{fig:acc_straggler_emnist_pdpe_all}}%
  \\
  \subfloat[\centering Total accuracy as a function of wall clock training time.]{{\includegraphics[width=0.95\linewidth]{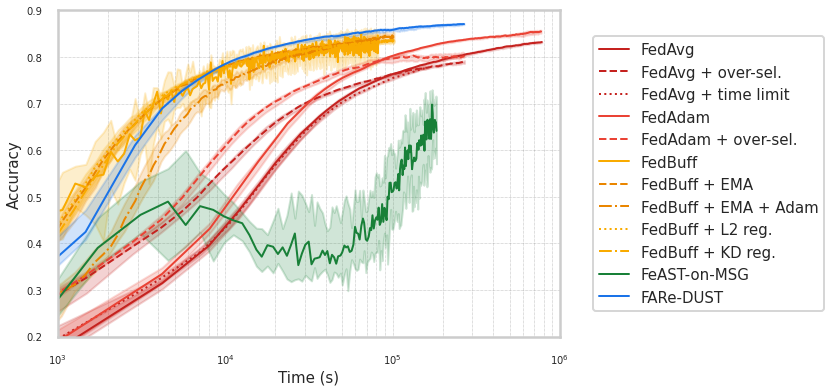}} \label{fig:acc_total_emnist_pdpe_all}}%
  \caption{Straggler accuracy (\ref{fig:acc_straggler_emnist_pdpe_all}) and total accuracy (\ref{fig:acc_total_emnist_pdpe_all}) as a function of wall clock training time for EMNIST with the \pdpe{} client latency scenario. Median and 90\% confidence interval values from 10 trials are illustrated with lines and bands, respectively.}%
  \label{fig:combined_acc_emnist_pdpe_all}%
\end{figure}

\begin{figure}%
  \centering
  \subfloat[\centering Straggler accuracy as a function of total accuracy.]{{\includegraphics[width=0.95\linewidth]{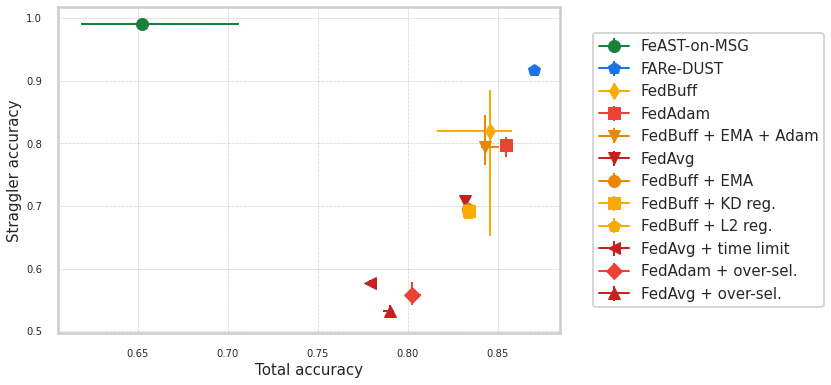}} \label{fig:2d_acc_emnist_pdpe_all}}%
  \\
  \subfloat[\centering Straggler accuracy as a function of total training time.]{{\includegraphics[width=0.95\linewidth]{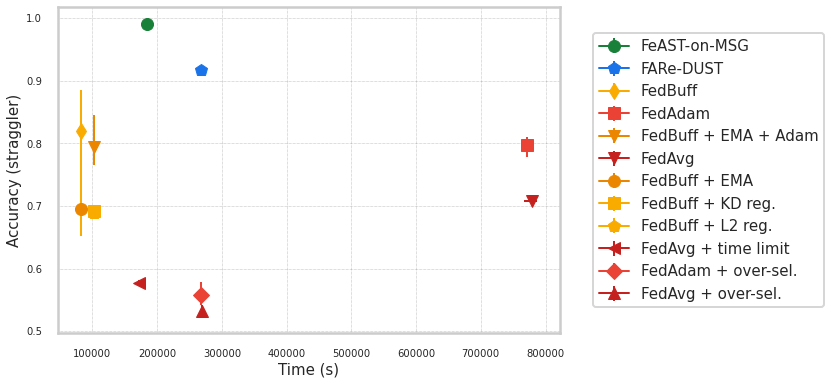}} \label{fig:2d_time_emnist_pdpe_all}}%
  \caption{Final straggler accuracy as a function of total accuracy (\ref{fig:2d_acc_emnist_pdpe_all}) and as a function of total wall clock training time (\ref{fig:2d_time_emnist_pdpe_all}) for EMNIST with the \pdpe{} client latency scenario. Median and 90\% confidence interval values from 10 trials are illustrated with points and error bars, respectively.}%
  \label{fig:combined_2d_emnist_pdpe_all}%
\end{figure}

\begin{figure}%
  \centering
  \subfloat[\centering Straggler accuracy as a function of wall clock training time.]{{\includegraphics[width=0.95\linewidth]{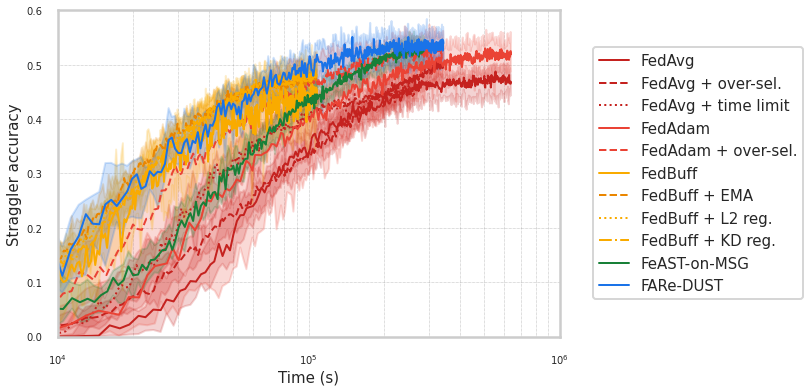}} \label{fig:acc_straggler_cifar_pe_all}}%
  \\
  \subfloat[\centering Total accuracy as a function of wall clock training time.]{{\includegraphics[width=0.95\linewidth]{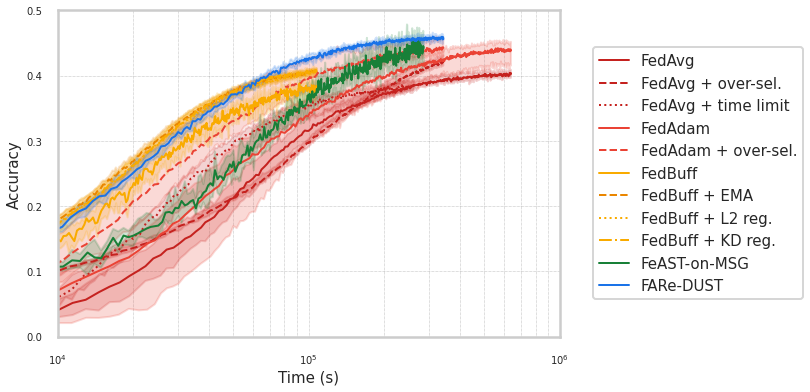}} \label{fig:acc_total_cifar_pe_all}}%
  \caption{Straggler accuracy (\ref{fig:acc_straggler_cifar_pe_all}) and total accuracy (\ref{fig:acc_total_cifar_pe_all}) as a function of wall clock training time for CIFAR-100 with the \pe{} client latency scenario. Median and 90\% confidence interval values from 10 trials are illustrated with lines and bands, respectively.}%
  \label{fig:combined_acc_cifar_pe_all}%
\end{figure}

\begin{figure}%
  \centering
  \subfloat[\centering Straggler accuracy as a function of total accuracy.]{{\includegraphics[width=0.95\linewidth]{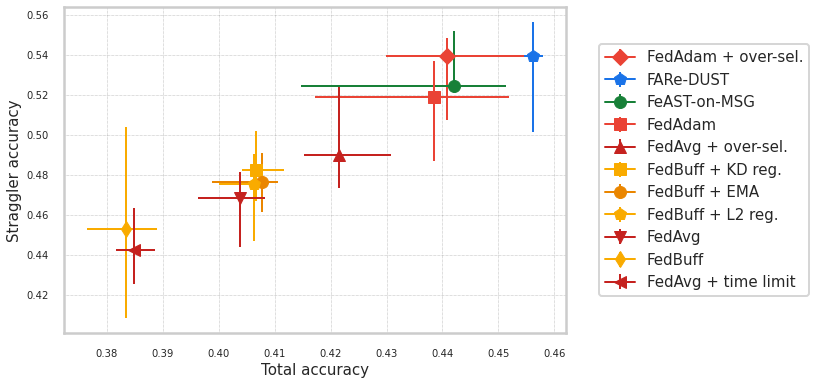}} \label{fig:2d_acc_cifar_pe_all}}%
  \\
  \subfloat[\centering Straggler accuracy as a function of total training time.]{{\includegraphics[width=0.95\linewidth]{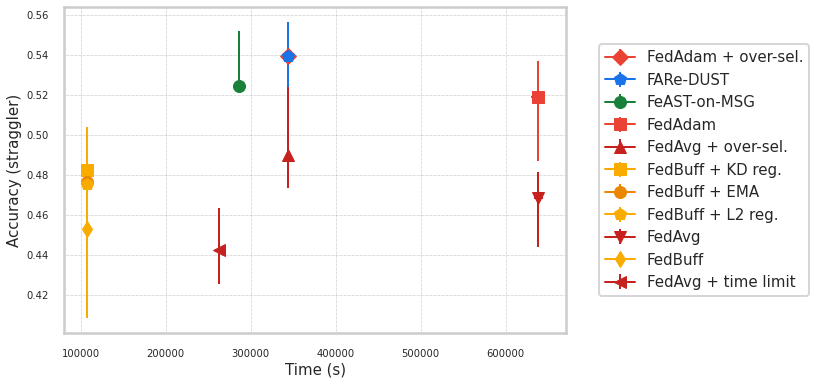}} \label{fig:2d_time_cifar_pe_all}}%
  \caption{Final straggler accuracy as a function of total accuracy (\ref{fig:2d_acc_cifar_pe_all}) and as a function of total wall clock training time (\ref{fig:2d_time_cifar_pe_all}) for CIFAR-100 with the \pe{} client latency scenario. Median and 90\% confidence interval values from 10 trials are illustrated with points and error bars, respectively.}%
  \label{fig:combined_2d_cifar_pe_all}%
\end{figure}

\begin{figure}%
  \centering
  \subfloat[\centering Straggler accuracy as a function of wall clock training time.]{{\includegraphics[width=0.95\linewidth]{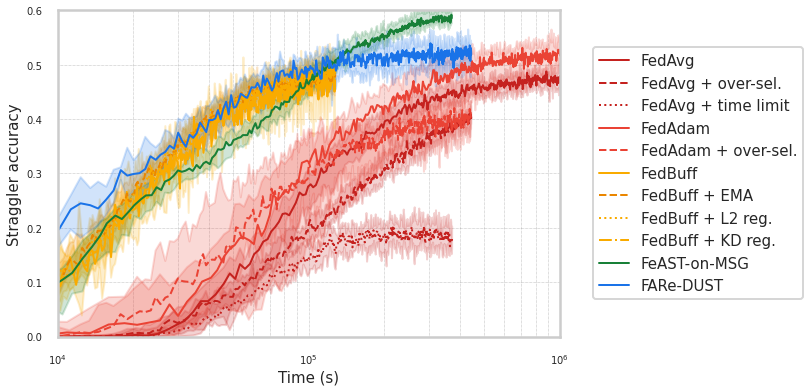}} \label{fig:acc_straggler_cifar_pdpe_all}}%
  \\
  \subfloat[\centering Total accuracy as a function of wall clock training time.]{{\includegraphics[width=0.95\linewidth]{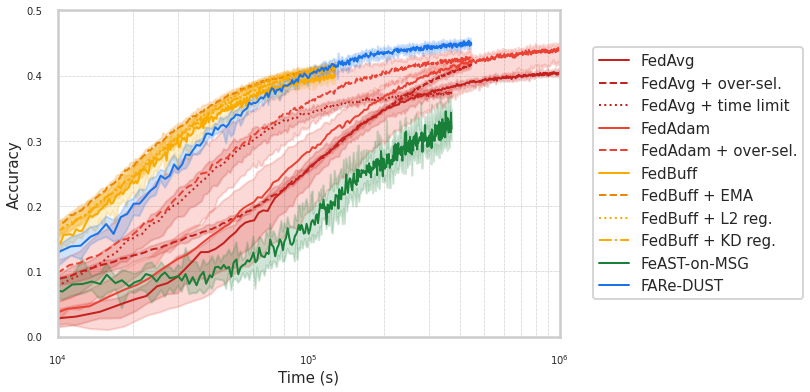}} \label{fig:acc_total_cifar_pdpe_all}}%
  \caption{Straggler accuracy (\ref{fig:acc_straggler_cifar_pdpe_all}) and total accuracy (\ref{fig:acc_total_cifar_pdpe_all}) as a function of wall clock training time for CIFAR-100 with the \pdpe{} client latency scenario. Median and 90\% confidence interval values from 10 trials are illustrated with lines and bands, respectively.}%
  \label{fig:combined_acc_cifar_pdpe_all}%
\end{figure}

\begin{figure}%
  \centering
  \subfloat[\centering Straggler accuracy as a function of total accuracy.]{{\includegraphics[width=0.95\linewidth]{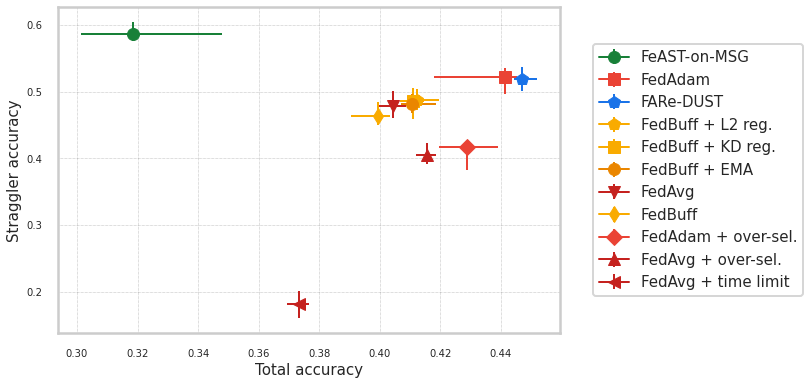}} \label{fig:2d_acc_cifar_pdpe_all}}%
  \\
  \subfloat[\centering Straggler accuracy as a function of total training time.]{{\includegraphics[width=0.95\linewidth]{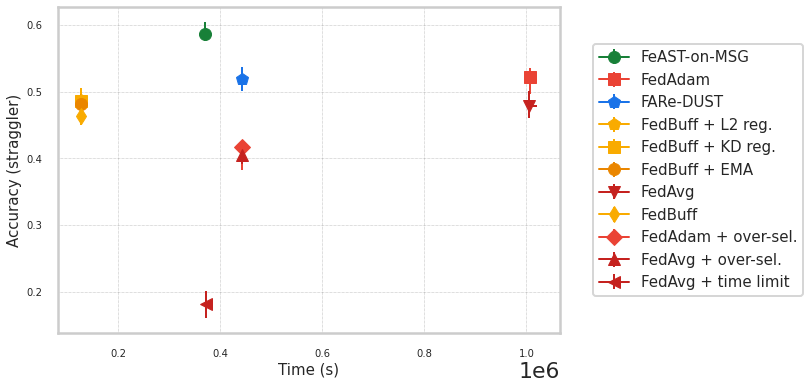}} \label{fig:2d_time_cifar_pdpe_all}}%
  \caption{Final straggler accuracy as a function of total accuracy (\ref{fig:2d_acc_cifar_pdpe_all}) and as a function of total wall clock training time (\ref{fig:2d_time_cifar_pdpe_all}) for CIFAR-100 with the \pdpe{} client latency scenario. Median and 90\% confidence interval values from 10 trials are illustrated with points and error bars, respectively.}%
  \label{fig:combined_2d_cifar_pdpe_all}%
\end{figure}

\begin{figure}%
  \centering
  \subfloat[\centering Straggler accuracy as a function of wall clock training time.]{{\includegraphics[width=0.95\linewidth]{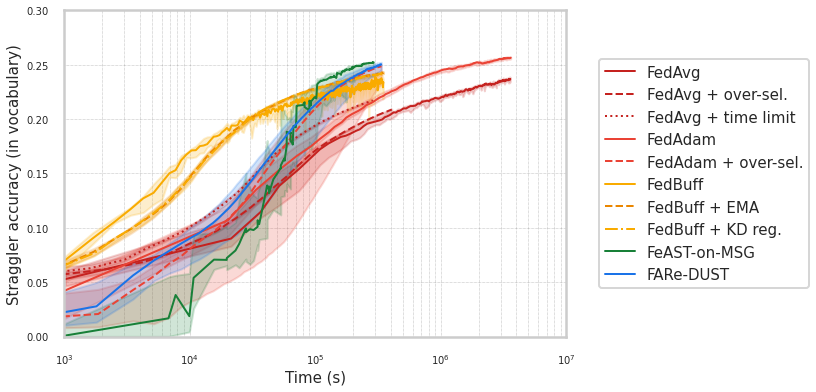}} \label{fig:acc_straggler_so_pe_all}}%
  \\
  \subfloat[\centering Total accuracy as a function of wall clock training time.]{{\includegraphics[width=0.95\linewidth]{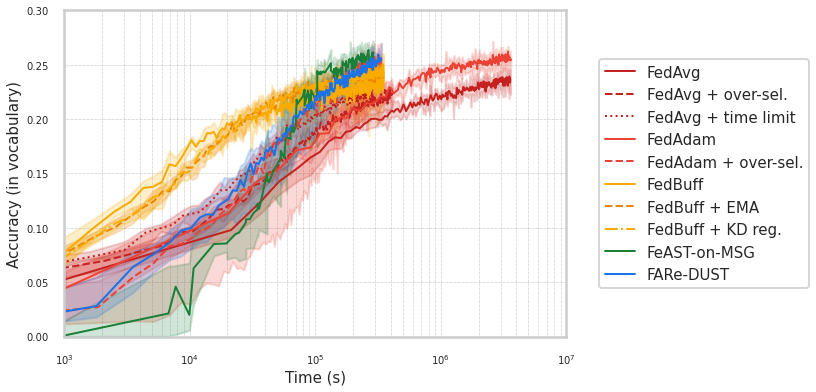}} \label{fig:acc_total_so_pe_all}}%
  \caption{Straggler accuracy (\ref{fig:acc_straggler_so_pe_all}) and total accuracy (\ref{fig:acc_total_so_pe_all}) as a function of wall clock training time for StackOverflow with the \pe{} client latency scenario. Median and 90\% confidence interval values from 10 trials are illustrated with lines and bands, respectively.}%
  \label{fig:combined_acc_so_pe_all}%
\end{figure}

\begin{figure}%
  \centering
  \subfloat[\centering Straggler accuracy as a function of total accuracy.]{{\includegraphics[width=0.95\linewidth]{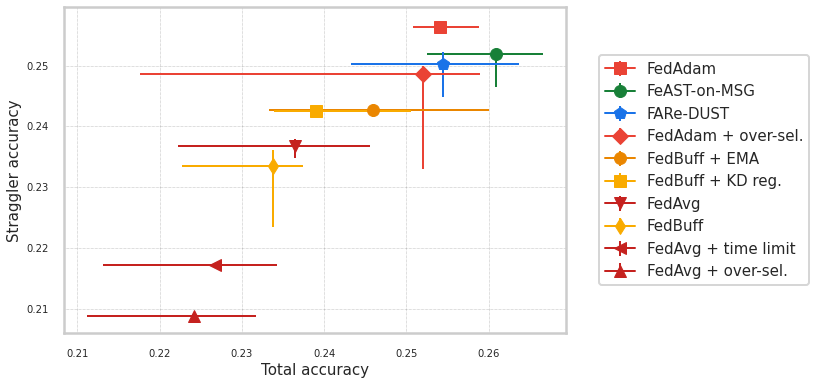}} \label{fig:2d_acc_so_pe_all}}%
  \\
  \subfloat[\centering Straggler accuracy as a function of total training time.]{{\includegraphics[width=0.95\linewidth]{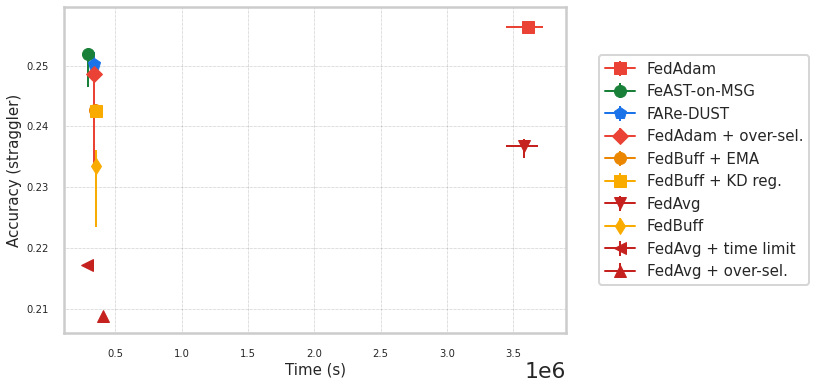}} \label{fig:2d_time_so_pe_all}}%
  \caption{Final straggler accuracy as a function of total accuracy (\ref{fig:2d_acc_so_pe_all}) and as a function of total wall clock training time (\ref{fig:2d_time_so_pe_all}) for StackOverflow with the \pe{} client latency scenario. Median and 90\% confidence interval values from 10 trials are illustrated with points and error bars, respectively.}%
  \label{fig:combined_2d_so_pe_all}%
\end{figure}

\begin{figure}%
  \centering
  \subfloat[\centering Straggler accuracy as a function of wall clock training time.]{{\includegraphics[width=0.95\linewidth]{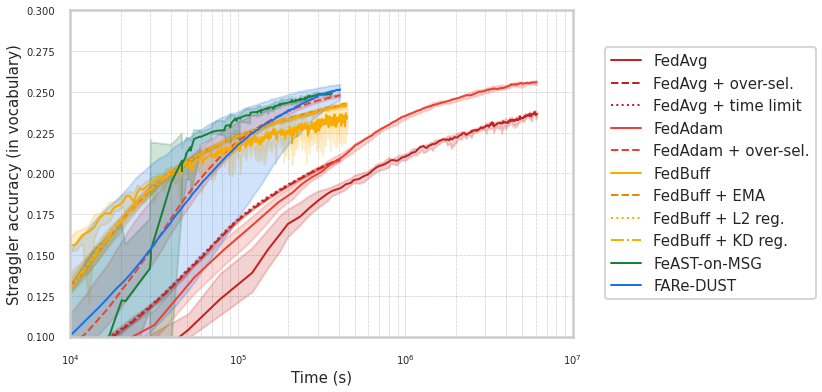}} \label{fig:acc_straggler_so_pdpe_all}}%
  \\
  \subfloat[\centering Total accuracy as a function of wall clock training time.]{{\includegraphics[width=0.95\linewidth]{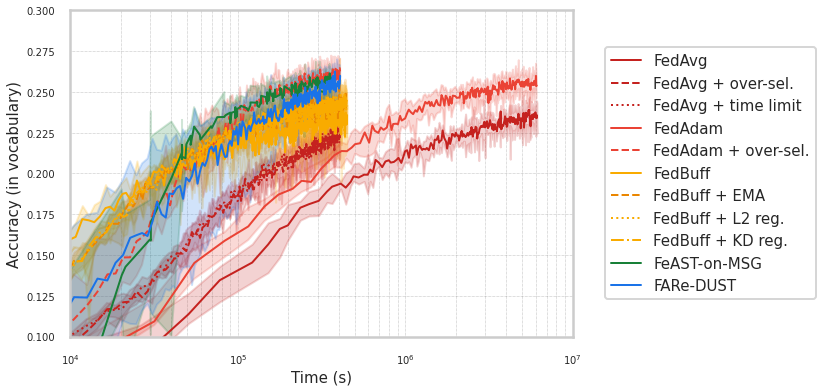}} \label{fig:acc_total_so_pdpe_all}}%
  \caption{Straggler accuracy (\ref{fig:acc_straggler_so_pdpe_all}) and total accuracy (\ref{fig:acc_total_so_pdpe_all}) as a function of wall clock training time for StackOverflow with the \pdpe{} client latency scenario. Median and 90\% confidence interval values from 10 trials are illustrated with lines and bands, respectively.}%
  \label{fig:combined_acc_so_pdpe_all}%
\end{figure}

\begin{figure}%
  \centering
  \subfloat[\centering Straggler accuracy as a function of total accuracy.]{{\includegraphics[width=0.95\linewidth]{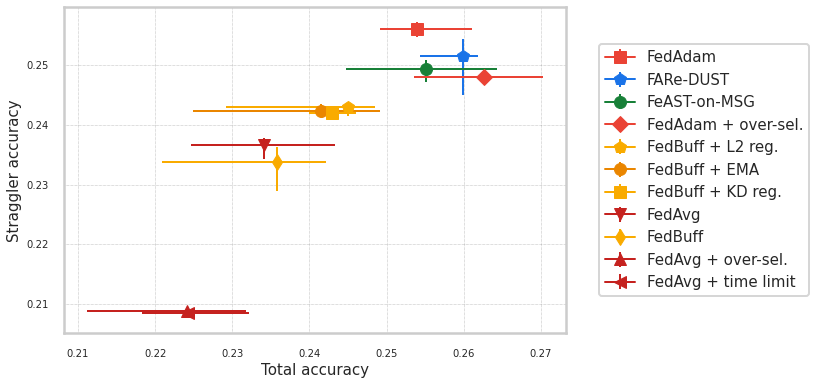}} \label{fig:2d_acc_so_pdpe_all}}%
  \\
  \subfloat[\centering Straggler accuracy as a function of total training time.]{{\includegraphics[width=0.95\linewidth]{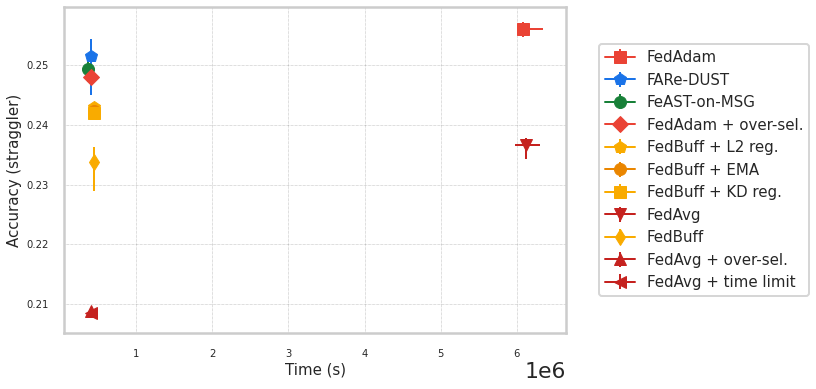}} \label{fig:2d_time_so_pdpe_all}}%
  \caption{Final straggler accuracy as a function of total accuracy (\ref{fig:2d_acc_so_pdpe_all}) and as a function of total wall clock training time (\ref{fig:2d_time_so_pdpe_all}) for StackOverflow with the \pdpe{} client latency scenario. Median and 90\% confidence interval values from 10 trials are illustrated with points and error bars, respectively.}%
  \label{fig:combined_2d_so_pdpe_all}%
\end{figure}

\begin{itemize}
  \item Results for EMNIST with the \pe{} client latency scenario.
  \begin{itemize}
    \item Table~\ref{table:emnist_pe_full}: summary of the final evaluation metrics, with algorithms sorted in descending order of straggler accuracy.
    \item Figure~\ref{fig:acc_straggler_emnist_pe_all}: straggler accuracy as a function of wall clock training time.
    \item Figure~\ref{fig:acc_total_emnist_pe_all}: total accuracy as a function of wall clock training time.
    \item Figure~\ref{fig:2d_acc_emnist_pe_all}: the final straggler accuracy as a function of the final total accuracy.
    \item Figure~\ref{fig:2d_time_emnist_pe_all}: the final straggler accuracy as a function of total wall clock training time.
  \end{itemize}
  \item Results for EMNIST with the \pdpe{} client latency scenario.
  \begin{itemize}
    \item Table~\ref{table:emnist_pdpe_full}: summary of the final evaluation metrics, with algorithms sorted in descending order of straggler accuracy.
    \item Figure~\ref{fig:acc_straggler_emnist_pdpe_all}: straggler accuracy as a function of wall clock training time.
    \item Figure~\ref{fig:acc_total_emnist_pdpe_all}: total accuracy as a function of wall clock training time.
    \item Figure~\ref{fig:2d_acc_emnist_pdpe_all}: the final straggler accuracy as a function of the final total accuracy.
    \item Figure~\ref{fig:2d_time_emnist_pdpe_all}: the final straggler accuracy as a function of total wall clock training time.
  \end{itemize}
  \item Results for CIFAR-100 with the \pe{} client latency scenario.
  \begin{itemize}
    \item Table~\ref{table:cifar_pe_full}: summary of the final evaluation metrics, with algorithms sorted in descending order of straggler accuracy.
    \item Figure~\ref{fig:acc_straggler_cifar_pe_all}: straggler accuracy as a function of wall clock training time.
    \item Figure~\ref{fig:acc_total_cifar_pe_all}: total accuracy as a function of wall clock training time.
    \item Figure~\ref{fig:2d_acc_cifar_pe_all}: the final straggler accuracy as a function of the final total accuracy.
    \item Figure~\ref{fig:2d_time_cifar_pe_all}: the final straggler accuracy as a function of total wall clock training time.
  \end{itemize}
  \item Results for CIFAR-100 with the \pdpe{} client latency scenario.
  \begin{itemize}
    \item Table~\ref{table:cifar_pdpe_full}: summary of the final evaluation metrics, with algorithms sorted in descending order of straggler accuracy.
    \item Figure~\ref{fig:acc_straggler_cifar_pdpe_all}: straggler accuracy as a function of wall clock training time.
    \item Figure~\ref{fig:acc_total_cifar_pdpe_all}: total accuracy as a function of wall clock training time.
    \item Figure~\ref{fig:2d_acc_cifar_pdpe_all}: the final straggler accuracy as a function of the final total accuracy.
    \item Figure~\ref{fig:2d_time_cifar_pdpe_all}: the final straggler accuracy as a function of total wall clock training time.
  \end{itemize}
  \item Results for StackOverflow with the \pe{} client latency scenario.
  \begin{itemize}
    \item Table~\ref{table:so_pe_full}: summary of the final evaluation metrics, with algorithms sorted in descending order of straggler accuracy.
    \item Figure~\ref{fig:acc_straggler_so_pe_all}: straggler accuracy as a function of wall clock training time.
    \item Figure~\ref{fig:acc_total_so_pe_all}: total accuracy as a function of wall clock training time.
    \item Figure~\ref{fig:2d_acc_so_pe_all}: the final straggler accuracy as a function of the final total accuracy.
    \item Figure~\ref{fig:2d_time_so_pe_all}: the final straggler accuracy as a function of total wall clock training time.
  \end{itemize}
  \item Results for StackOverflow with the \pdpe{} client latency scenario.
  \begin{itemize}
    \item Table~\ref{table:so_pdpe_full}: summary of the final evaluation metrics, with algorithms sorted in descending order of straggler accuracy.
    \item Figure~\ref{fig:acc_straggler_so_pdpe_all}: straggler accuracy as a function of wall clock training time.
    \item Figure~\ref{fig:acc_total_so_pdpe_all}: total accuracy as a function of wall clock training time.
    \item Figure~\ref{fig:2d_acc_so_pdpe_all}: the final straggler accuracy as a function of the final total accuracy.
    \item Figure~\ref{fig:2d_time_so_pdpe_all}: the final straggler accuracy as a function of total wall clock training time.
  \end{itemize}
\end{itemize}


\begin{table}
  \caption{The straggler accuracy, total accuracy, and the median total training time for each algorithm with the EMNIST dataset and \pe{} client latency model. Median and 90\% CI values from 10 trials are quoted for each metric. Rows are sorted in descending order of straggler accuracy.}
  \label{table:emnist_pe_full}
  \vskip 0.15in
  \centering
  \begin{tabular}{lccccr}
    \toprule
    & \multicolumn{2}{c}{Straggler accuracy} & \multicolumn{2}{c}{Total accuracy} & \\
    \cmidrule(r){2-3} \cmidrule(r){4-5}
    Algorithm & Median & 95\% CI & Median & 95\% CI & Time [s] \\
    \midrule 
    FedAdam + over-sel.  & 81.3\% & [79.6\%, 82.2\%] & 85.4\% & [85.2\%, 85.7\%] & 221,268 \\
    FeAST-on-MSG         & 81.0\% & [78.5\%, 84.8\%] & 85.2\% & [84.4\%, 87.7\%] & 184,630 \\
    FedAdam              & 79.1\% & [76.6\%, 80.1\%] & 85.3\% & [84.9\%, 85.6\%] & 475,782 \\
    FedBuff + EMA + Adam & 77.8\% & [75.9\%, 84.4\%] & 84.2\% & [83.3\%, 84.9\%] &  90,534 \\
    FARe-DUST            & 77.5\% & [74.4\%, 78.6\%] & 85.3\% & [84.7\%, 85.4\%] & 221,166 \\
    FedAvg + over-sel.   & 72.3\% & [71.2\%, 73.1\%] & 83.3\% & [83.0\%, 83.4\%] & 222,279 \\
    FedAvg + time limit  & 71.8\% & [70.8\%, 72.5\%] & 83.2\% & [83.0\%, 83.3\%] & 173,012 \\
    FedAvg               & 71.1\% & [70.5\%, 71.6\%] & 83.2\% & [83.1\%, 83.3\%] & 479,538 \\
    FedBuff + L2 reg.    & 69.0\% & [67.7\%, 69.8\%] & 83.2\% & [82.8\%, 83.5\%] &  90,547 \\
    FedBuff + EMA        & 68.9\% & [67.2\%, 69.6\%] & 82.8\% & [82.3\%, 83.0\%] &  72,554 \\
    FedBuff + KD reg.    & 68.6\% & [67.6\%, 69.6\%] & 82.9\% & [82.5\%, 83.1\%] &  90,503 \\
    FedBuff              & 61.2\% & [52.5\%, 79.6\%] & 79.7\% & [77.3\%, 84.3\%] &  72,480 \\
    \bottomrule
  \end{tabular}
\end{table}

\begin{table}
  \caption{The straggler accuracy, total accuracy, and the total training time for each algorithm with the EMNIST dataset and \pdpe{} client latency model. Median and 90\% CI values from 10 trials are quoted for each metric. Rows are sorted in descending order of straggler accuracy.}
  \label{table:emnist_pdpe_full}
  \vskip 0.15in
  \centering
  \begin{tabular}{lccccr}
    \toprule
    & \multicolumn{2}{c}{Straggler accuracy} & \multicolumn{2}{c}{Total accuracy} &                   \\
    \cmidrule(r){2-3} \cmidrule(r){4-5}
    Algorithm & Median & 95\% CI & Median & 95\% CI & Time [s] \\
    \midrule 
    FeAST-on-MSG         & 99.2\% & [98.9\%, 99.4\%] & 65.2\% & [61.8\%, 70.6\%] & 184,565 \\
    FARe-DUST            & 91.7\% & [91.0\%, 92.5\%] & 87.0\% & [86.9\%, 87.2\%] & 267,541 \\
    FedBuff              & 82.0\% & [65.2\%, 88.5\%] & 84.6\% & [81.6\%, 85.8\%] &  81,798 \\
    FedAdam              & 79.8\% & [77.8\%, 81.1\%] & 85.4\% & [85.1\%, 85.7\%] & 771,556 \\
    FedBuff + EMA + Adam & 79.4\% & [76.5\%, 84.5\%] & 84.3\% & [84.1\%, 85.1\%] & 102,138 \\
    FedAvg               & 70.8\% & [70.3\%, 71.2\%] & 83.2\% & [83.0\%, 83.2\%] & 779,631 \\
    FedBuff + EMA        & 69.5\% & [68.4\%, 70.2\%] & 83.4\% & [83.1\%, 83.6\%] &  81,902 \\
    FedBuff + KD reg.    & 69.1\% & [68.3\%, 69.8\%] & 83.4\% & [83.2\%, 83.6\%] & 102,150 \\
    FedBuff + L2 reg.    & 68.9\% & [68.1\%, 70.0\%] & 83.4\% & [83.1\%, 83.6\%] & 102,115 \\
    FedAvg + time limit  & 57.8\% & [57.3\%, 58.2\%] & 77.9\% & [77.8\%, 78.0\%] & 171,958 \\
    FedAdam + over-sel.  & 55.8\% & [54.2\%, 57.8\%] & 80.2\% & [79.9\%, 80.8\%] & 267,422 \\
    FedAvg + over-sel.   & 53.2\% & [52.0\%, 54.2\%] & 79.0\% & [78.6\%, 79.2\%] & 269,295 \\
    \bottomrule
  \end{tabular}
\end{table}

\begin{table}
  \caption{The straggler accuracy, total accuracy, and the total training time for each algorithm with the CIFAR-100 dataset and \pe{} client latency model. Median and 90\% CI values from 10 trials are quoted for each metric. Rows are sorted in descending order of straggler accuracy.}
  \label{table:cifar_pe_full}
  \vskip 0.15in
  \centering
  \begin{tabular}{lccccr}
    \toprule
    & \multicolumn{2}{c}{Straggler accuracy} & \multicolumn{2}{c}{Total accuracy} & \\
    \cmidrule(r){2-3} \cmidrule(r){4-5}
    Algorithm & Median & 95\% CI & Median & 95\% CI & Time [s] \\
    \midrule 
    FedAdam + over-sel. & 54.0\% & [50.7\%, 54.9\%] & 44.1\% & [43.0\%, 45.5\%] & 343,556 \\
    FARe-DUST           & 54.0\% & [50.1\%, 55.7\%] & 45.6\% & [45.5\%, 45.8\%] & 343,502 \\
    FeAST-on-MSG        & 52.5\% & [52.3\%, 55.2\%] & 44.2\% & [41.5\%, 45.1\%] & 286,516 \\
    FedAdam             & 51.9\% & [48.7\%, 53.7\%] & 43.8\% & [41.7\%, 45.2\%] & 638,146 \\
    FedAvg + over-sel.  & 49.0\% & [47.3\%, 52.4\%] & 42.2\% & [41.5\%, 43.1\%] & 343,386 \\
    FedBuff + KD reg.   & 48.3\% & [46.7\%, 50.2\%] & 40.7\% & [40.4\%, 41.2\%] & 107,530 \\
    FedBuff + EMA       & 47.7\% & [46.1\%, 49.1\%] & 40.8\% & [39.9\%, 41.1\%] & 107,582 \\
    FedBuff + L2 reg.   & 47.6\% & [44.7\%, 49.0\%] & 40.6\% & [40.0\%, 40.8\%] & 107,456 \\
    FedAvg              & 46.9\% & [44.4\%, 48.1\%] & 40.4\% & [39.6\%, 40.8\%] & 638,038 \\
    FedBuff             & 45.3\% & [40.8\%, 50.4\%] & 38.3\% & [37.6\%, 38.9\%] & 107,582 \\
    FedAvg + time limit & 44.3\% & [42.6\%, 46.3\%] & 38.5\% & [38.2\%, 38.9\%] & 263,174 \\
    \bottomrule
  \end{tabular}
\end{table}

\begin{table}
  \caption{The straggler accuracy, total accuracy, and the total training time for each algorithm with the CIFAR-100 dataset and \pdpe{} client latency model. Median and 90\% CI values from 10 trials are quoted for each metric. Rows are sorted in descending order of straggler accuracy.}
  \label{table:cifar_pdpe_full}
  \vskip 0.15in
  \centering
  \begin{tabular}{lccccr}
    \toprule
    & \multicolumn{2}{c}{Straggler accuracy} & \multicolumn{2}{c}{Total accuracy} &                   \\
    \cmidrule(r){2-3} \cmidrule(r){4-5}
    Algorithm & Median & 95\% CI & Median & 95\% CI & Time [s] \\
    \midrule 
    FeAST-on-MSG        & 58.6\% & [57.6\%, 60.4\%] & 31.9\% & [30.1\%, 34.8\%] &   370,547 \\
    FedAdam             & 52.2\% & [49.7\%, 53.6\%] & 44.2\% & [41.8\%, 44.5\%] & 1,006,330 \\
    FARe-DUST           & 51.9\% & [50.1\%, 53.7\%] & 44.7\% & [44.5\%, 45.2\%] &   443,452 \\
    FedBuff + L2 reg.   & 48.8\% & [47.3\%, 50.4\%] & 41.2\% & [41.0\%, 42.0\%] &   127,336 \\
    FedBuff + KD reg.   & 48.6\% & [45.9\%, 50.5\%] & 41.1\% & [40.7\%, 41.5\%] &   127,375 \\
    FedBuff + EMA       & 48.2\% & [46.7\%, 49.6\%] & 41.1\% & [40.7\%, 41.9\%] &   127,448 \\
    FedAvg              & 47.9\% & [46.1\%, 50.1\%] & 40.4\% & [40.0\%, 40.9\%] & 1,004,752 \\
    FedBuff             & 46.4\% & [44.9\%, 48.4\%] & 40.0\% & [39.1\%, 40.3\%] &   127,208 \\
    FedAdam + over-sel. & 41.7\% & [38.3\%, 42.3\%] & 42.9\% & [41.9\%, 43.9\%] &   442,894 \\
    FedAvg + over-sel.  & 40.5\% & [39.2\%, 42.3\%] & 41.6\% & [41.2\%, 41.9\%] &   443,618 \\
    FedAvg + time limit & 18.3\% & [16.1\%, 20.2\%] & 37.3\% & [37.0\%, 37.7\%] &   372,209 \\
    \bottomrule
  \end{tabular}
\end{table}

\begin{table}
  \caption{The straggler accuracy, total accuracy, and the total training time for each algorithm with the StackOverflow dataset and \pe{} client latency model. Median and 90\% CI values from 10 trials are quoted for each metric. Rows are sorted in descending order of straggler accuracy.}
  \label{table:so_pe_full}
  \vskip 0.15in
  \centering
  \begin{tabular}{lccccr}
    \toprule
    & \multicolumn{2}{c}{Straggler accuracy} & \multicolumn{2}{c}{Total accuracy} & \\
    \cmidrule(r){2-3} \cmidrule(r){4-5}
    Algorithm & Median & 95\% CI & Median & 95\% CI & Time [s] \\
    \midrule 
    FedAdam             & 25.6\% & [25.5\%, 25.7\%] & 25.4\% & [25.1\%, 25.9\%] & 3,612,004 \\
    FeAST-on-MSG        & 25.2\% & [24.6\%, 25.3\%] & 26.1\% & [25.2\%, 26.7\%] &   292,006 \\
    FARe-DUST           & 25.0\% & [24.5\%, 25.2\%] & 25.4\% & [24.3\%, 26.4\%] &   335,744 \\
    FedAdam + over-sel. & 24.9\% & [23.3\%, 25.0\%] & 25.2\% & [21.8\%, 25.9\%] &   336,140 \\
    FedBuff + EMA       & 24.3\% & [24.2\%, 24.3\%] & 24.6\% & [23.3\%, 26.0\%] &   349,178 \\
    FedBuff + KD reg.   & 24.2\% & [24.2\%, 24.3\%] & 23.9\% & [23.4\%, 25.1\%] &   350,286 \\
    FedAvg              & 23.7\% & [23.5\%, 23.8\%] & 23.6\% & [22.2\%, 24.6\%] & 3,584,260 \\
    FedBuff             & 23.3\% & [22.3\%, 23.6\%] & 23.4\% & [22.3\%, 23.7\%] &   349,257 \\
    FedAvg + time limit & 21.7\% & [21.7\%, 21.7\%] & 22.7\% & [21.3\%, 23.4\%] &   287,952 \\
    FedAvg + over-sel.  & 20.9\% & [20.8\%, 20.9\%] & 22.4\% & [21.1\%, 23.2\%] &   407,588 \\
    \bottomrule
  \end{tabular}
\end{table}

\begin{table}
  \caption{The straggler accuracy, total accuracy, and the total training time for each algorithm with the StackOverflow dataset and \pdpe{} client latency model. Median and 90\% CI values from 10 trials are quoted for each metric. Rows are sorted in descending order of straggler accuracy.}
  \label{table:so_pdpe_full}
  \vskip 0.15in
  \centering
  \begin{tabular}{lccccr}
    \toprule
    & \multicolumn{2}{c}{Straggler accuracy} & \multicolumn{2}{c}{Total accuracy} &                   \\
    \cmidrule(r){2-3} \cmidrule(r){4-5}
    Algorithm & Median & 95\% CI & Median & 95\% CI & Time [s] \\
    \midrule 
    FedAdam             & 25.6\% & [25.5\%, 25.7\%] & 25.4\% & [24.9\%, 26.1\%] & 6,086,590 \\
    FARe-DUST           & 25.1\% & [24.5\%, 25.4\%] & 26.0\% & [25.4\%, 26.2\%] &   408,276 \\
    FeAST-on-MSG        & 24.9\% & [24.7\%, 25.1\%] & 25.5\% & [24.5\%, 26.4\%] &   364,537 \\
    FedAdam + over-sel. & 24.8\% & [24.7\%, 24.9\%] & 26.3\% & [25.4\%, 27.0\%] &   408,178 \\
    FedBuff + L2 reg.   & 24.3\% & [24.1\%, 24.4\%] & 24.5\% & [22.9\%, 24.9\%] &   447,008 \\
    FedBuff + EMA       & 24.2\% & [24.2\%, 24.3\%] & 24.1\% & [22.5\%, 24.9\%] &   443,355 \\
    FedBuff + KD reg.   & 24.2\% & [24.1\%, 24.2\%] & 24.3\% & [24.0\%, 24.6\%] &   445,861 \\
    FedAvg              & 23.7\% & [23.4\%, 23.8\%] & 23.4\% & [22.5\%, 24.3\%] & 6,119,154 \\
    FedBuff             & 23.4\% & [22.9\%, 23.6\%] & 23.6\% & [22.1\%, 24.2\%] &   449,020 \\
    FedAvg + over-sel.  & 20.9\% & [20.8\%, 20.9\%] & 22.4\% & [21.1\%, 23.2\%] &   407,588 \\
    FedAvg + time limit & 20.9\% & [20.8\%, 20.9\%] & 22.4\% & [21.8\%, 23.2\%] &   405,116 \\
    \bottomrule
  \end{tabular}
\end{table}

\end{document}